\documentclass{article} 
\usepackage{iclr2027_conference,times}

\usepackage{amsmath,amsfonts,bm}

\def\eqref#1{equation~\ref{#1}}

\def\1{\bm{1}}

\DeclareMathAlphabet{\mathsfit}{\encodingdefault}{\sfdefault}{m}{sl}
\SetMathAlphabet{\mathsfit}{bold}{\encodingdefault}{\sfdefault}{bx}{n}

\newcommand{\Ljab}{\mathcal{L}_{\mathrm{JAB}}}  

\usepackage{hyperref}
\hypersetup{hidelinks}  
\usepackage{url}

\usepackage{amsmath,amssymb,amsfonts}
\usepackage{graphicx}
\usepackage{booktabs}
\usepackage{array}
\usepackage{multirow}
\usepackage{bm}
\usepackage{enumitem}
\graphicspath{{figures/}}

\title{\Large From Attention Sensitivity to Layer Role: Revisiting Mixed-Precision Quantization of Transformers}

\author{\end{tabular}\hspace*{-4\tabcolsep}\begin{tabular}[t]{c}
  \textbf{Nafiseh HosseinpourFardi}$^{1,*}$ \quad
  \textbf{Negar Alihadi}$^{1,*}$ \\[0.25em]
  \textbf{Mahmoudreza Babaei}$^{2,3,4}$ \quad
  \textbf{Milad Hosseini}$^{5}$ \quad
  \textbf{Adrian Weller}$^{6}$ \\[0.8em]
  \normalfont\normalsize
  $^{1}$Sharif University of Technology, Tehran, Iran \\
  $^{2}$GISMA University of Applied Sciences, Potsdam, Germany \\
  $^{3}$BRAINS, Brandenburg Research Center for Applied Intelligent Systems, Potsdam, Germany \\
  $^{4}$Max Planck Institute for Security and Privacy, Bochum, Germany \\
  $^{5}$Computer Vision Center (CVC), Barcelona, Spain \\
  $^{6}$University of Cambridge, United Kingdom \\[0.5em]
  $^{*}$Equal contribution.
}

\iclrfinalcopy  
\begin{document}

\maketitle
\lhead{}\renewcommand{\headrulewidth}{0pt}  

\begin{abstract}
Most post-training quantization pipelines keep each quantized weight matrix close to its pretrained
counterpart, one matrix at a time, and stop there. Whether that proxy tracks what the attention
block actually computes, or how errors in the Q, K and V projections compound once they meet inside
the softmax, is rarely checked. The obvious fix is to write the objective on the attention output
itself, over all three projections at once, and to use that same objective wherever the pipeline
needs a signal. That is what \textbf{JAB} does. It defines one scalar loss over the joint Q, K, V
weights of a block, evaluated against the block's real causally-masked attention output, and uses
it both to fit the quantized weights (GPTQ warm start, then STE with learnable scales) and to score
the block for a multiple-choice knapsack allocation.

On attention-only quantization of Mistral-7B this pays off: at 3 bits JAB recovers $77$--$90\%$ of
the gap between uniform GPTQ and full precision, and its sensitivity estimate tracks an oracle
costing 73 forward passes to within a fraction of a perplexity point. Once MLP layers enter the allocation, though, JAB is beaten by a role-aware offset rule that
needs no sensitivity estimate at all, both on GPT-2 with QKV and MLP quantized together and on the
full Mistral-7B model. With a 3-bit floor, this rule quantizes $96.4\%$ of Mistral-7B's
weights to 4.5 bits per parameter at $6.933$ perplexity --- within $4.4\%$ of full precision
($6.643$) at $3.56\times$ compression, 14\,GB to 3.93\,GB --- against $7.158$ for JAB at the same
budget, and to 3.5 bits at $7.875$ ($4.57\times$). Whichever matrix a weight belongs
to inside its block matters more than any sensitivity estimate we managed to compute.

Two things came out sideways. Block-local reconstruction turned out to be an unreliable proxy for
end-to-end perplexity: in one run a $4.6\times$ improvement in a block's own objective came with a
$32\times$ increase in perplexity, which is why every allocation in this paper is validated
end-to-end rather than on local objectives alone. And on attention-only quantization, fine-tuning
consistently moved weights farther from their pretrained values while pulling attention outputs
closer, with net gains. Post-training seems to recover attention behavior, not weights.

\end{abstract}
\section{Introduction}
\label{sec:intro}

Large language models (LLMs) face severe memory bottlenecks during inference, and quantization
mitigates these costs by mapping real-valued weights to discrete integers \citep{mixedPrecision}.
Because quantization-aware training (QAT) remains prohibitively resource-intensive, post-training
quantization (PTQ) has become the dominant paradigm for efficient compression
\citep{enhancing_ultra_low}.

Early quantization methods relied on the inverse Hessian matrix of the Optimal Brain Surgeon (OBS)
algorithm, which proved computationally intractable for modern deep networks
\citep{hassibi1993obs,awp}. GPTQ resolved this by processing weights in blocks via Cholesky
decomposition, which made it possible to compress 175-billion-parameter LLMs such as GPT-3, models
that are prohibitively expensive to deploy in standard FP16 precision, to 3 or 4 bits efficiently.
Building on OBS, GPTQ minimizes a layer-wise weight-reconstruction loss; this local weight-space
objective can, however, fail to capture end-to-end Transformer fidelity \citep{frantar2023gptq}.

Pushing below 4 bits exposes structural vulnerabilities to heavy-tailed activation outliers.
Outlier-Aware Weight Quantization (OWQ) addresses this by preserving a subset of sensitive weights in
higher precision \citep{owq}. To automate variable bit-width allocation under strict budgets,
HAWQ-V2 sets bit widths from the average Hessian trace of the task loss, which serves as a measure of
parameter sensitivity \citep{dong2020hawqv2}, and the resulting assignment is optimized with
Multiple-Choice Knapsack Problem (MCKP) solvers \citep{sinha1979mckp,q_strata}. HAQ learns the
allocation instead \citep{wang2019haq}.

Moving beyond independent layer-wise optimization, BRECQ introduced block-wise reconstruction to
capture cross-layer dependencies \citep{li2021brecq}. Attention-Aware PTQ (APTQ) narrows the gap for
attention layers by folding attention information --- the gradients of the non-linear softmax ---
into its sensitivity estimate to guide optimization, but that criterion is still a weight-space
proxy, not the reconstruction objective itself \citep{guan2024aptq}.

In most PTQ pipelines, then, the objective that reconstructs the quantized weights is not the
criterion that sets their bit widths (Appendix~\ref{app:related}). This disconnect between the
sensitivity-estimation criterion and the actual reconstruction loss, which we term the
\emph{proxy gap}, motivates the Joint Attention-Based (JAB) framework: we set out to test whether one
objective could do both jobs, produce the weights and set the budget.

\textbf{JAB} (Joint Attention-Based allocation) uses a single loss for both steps. We define $\Ljab$
over the concatenated $[W_Q \mid W_K \mid W_V]$ parameters of a block and evaluate it against that
block's own causally-masked attention output and attention map. We then use it twice. Minimizing it
directly produces the weights: a GPTQ warm start, then straight-through fine-tuning with learnable
scales. Differentiating it twice, through Hutchinson probes of the Hessian trace, scores the block for
an MCKP allocation. We ask whether the same machinery transfers once allocation extends past
attention to the MLP and to the whole model.

\paragraph{Findings}
Attention-only and heterogeneous full-model quantization separate cleanly. On attention-only
Mistral-7B, JAB closes most of the gap between uniform GPTQ and full precision at 3 bits, recovering
$77$--$90\%$ of it across all four calibration/evaluation corpus pairings. Its Hessian-trace
criterion tracks the end-to-end oracle closely too, and at a small fraction of the model evaluations.
The same advantage does not carry over to the MLP or the full model. On GPT-2's MLP the oracle
allocates bits mainly by matrix role rather than depth, and the Hessian criterion does not pick that
structure up. On the full Mistral-7B model a seven-parameter role offset rule reaches lower
perplexity than JAB at the same parameter-weighted budget, with no sensitivity estimation at all.
With a 3-bit floor on every unit, the same rule reaches $6.933$ perplexity at 4.5 bits per
parameter, within $4.4\%$ of full precision ($6.643$) at $3.56\times$ compression, and $7.875$ at
3.5 bits ($4.57\times$).
Two further results bear on block-local refinement. Improving a block's local reconstruction
objective does not guarantee better end-to-end quality: in one experiment a $4.6\times$ reduction in
the local objective came with a $32\times$ increase in end-to-end perplexity. And on attention-only
quantization, fine-tuning moves the weights farther from their floating-point values while making
the attention outputs more accurate, which points to functional recovery mattering more than
weight-space fidelity.

\paragraph{Contributions}
\begin{itemize}[leftmargin=*,itemsep=1pt,topsep=2pt]
  \item \textbf{Joint objective for reconstruction and allocation.} We introduce JAB: a single joint
    QKV attention loss used for both quantized-weight refinement and sensitivity-based bit
    allocation. We also separate the joint from the per-matrix treatment of QKV. The two agree in
    expectation, but the per-matrix form can underestimate the variability of the true joint loss
    (Section~\ref{sec:joint}).
  \item \textbf{Evidence for role-dependent allocation.} MLP and full-model experiments show that
    matrix role can provide a stronger allocation signal than the curvature criterion JAB uses, which
    motivates a Hessian-free role-based prior (Sections~\ref{sec:m-mlp} and~\ref{sec:m-full}).
  \item \textbf{Local-to-global failure mode.} We isolate a case where improving a block-local
    reconstruction objective substantially worsens end-to-end perplexity. Local refinement
    objectives alone do not guarantee global quality (Section~\ref{sec:m-proxygap}).
  \item \textbf{Attention-space interpretation of refinement.} We introduce the amplification factor
    $\rho = \varepsilon^A/\varepsilon^W$ to separate weight fidelity from functional fidelity.
    Attention-aware refinement can improve the latter while moving the weights farther from their
    floating-point values (Section~\ref{sec:m-qkv}).
\end{itemize}

A parallel line of PTQ work improves the quantizer itself rather than the allocation, through
activation-aware scaling \citep{lin2024awq,xiao2023smoothquant}, learned clipping \citep{shao2024omniquant}, or incoherence processing \citep{tseng2024quip}. These are orthogonal to
bit allocation and compose with it; we hold the quantizer fixed and vary only how bits are assigned.

\section{Method}
\label{sec:method}

\begin{figure}[t]
\centering
\includegraphics[width=0.56\textwidth]{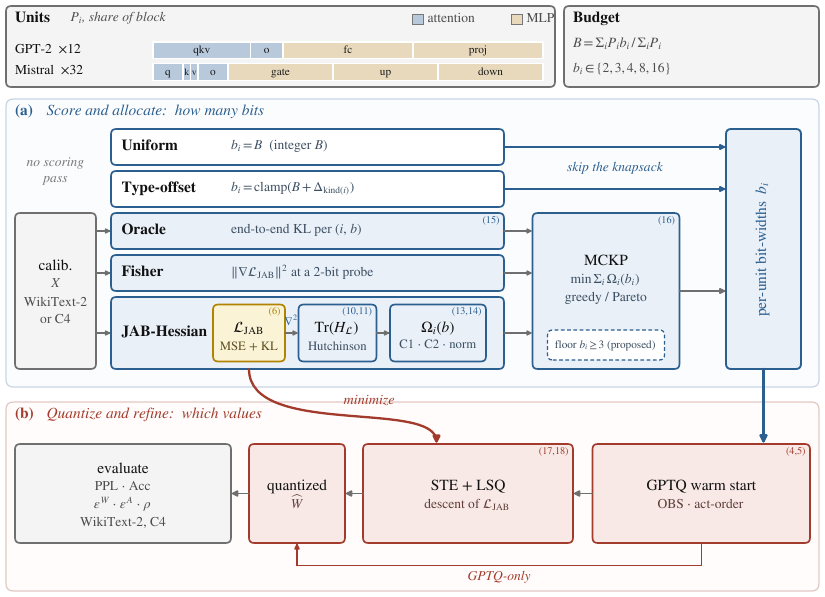}
\caption{Overview of the proposed mixed-precision PTQ pipeline. The method separates the process
into two stages. \textbf{(a) Bit allocation:} candidate sensitivity criteria determine how many bits
each unit receives under a parameter-weighted budget.  \textbf{(b) Quantization and refinement:} GPTQ provides the initial
quantized weights, after which the joint attention loss $\Ljab$ is minimized through the quantizer
using STE${+}$LSQ.}
\label{fig:pipeline}
\end{figure}

\subsection{Layer-wise PTQ}
\label{sec:gptq}
Let $X \in \mathbb{R}^{B \times T \times d}$ be a calibration input to one block and
$y = xW$ a sublayer with $W \in \mathbb{R}^{d_{\text{in}} \times d_{\text{out}}}$; the QKV
projection fuses $W = [W_Q \mid W_K \mid W_V]$, and attention is
$A(X) = \operatorname{softmax}\big(QK^\top/\sqrt{d_{\text{head}}} + M\big) V$ with causal mask $M$.
GPTQ minimizes the local surrogate
\begin{equation}
\mathcal{L}(\widehat{W}) = \textstyle\sum_{x \in X} \| xW - x\widehat{W} \|_2^2 ,
\qquad H = 2 X^\top X ,
\label{eq:local_loss}
\end{equation}
placing each scalar on a symmetric grid with a per-(row, group) scale $s$,
\begin{equation}
Q(w) = \operatorname{clip}\!\big(\operatorname{round}(w/s), -q_{\max}, q_{\max}\big)\cdot s,
\qquad q_{\max} = 2^{b-1}\!-\!1 ,
\label{eq:quant}
\end{equation}
and compensating not-yet-quantized columns with the OBS update (Appendix~\ref{app:gptq}).
$H$ is accumulated by a forward hook on the model in its \emph{current} state, so block $i$'s
Hessian already reflects blocks $0,\dots,i{-}1$ being quantized. The solve runs in float64; in
float32 the column loop's repeated division by the inverse-Hessian diagonal makes perplexity
non-monotone in the bit-width.

\subsection{The joint attention loss}
\label{sec:loss}
Let $A(X), P(X)$ be the attention output and post-softmax map of the unmodified block, and
$\widehat{A}(X;w), Q(X;w)$ those of a differentiable re-implementation evaluated at the quantized
$w = [\operatorname{vec}(\widehat{W}_Q); \operatorname{vec}(\widehat{W}_K);
\operatorname{vec}(\widehat{W}_V)]$. The joint loss is
\begin{equation}
\Ljab(w) = \big\| A(X) - \widehat{A}(X; w) \big\|_F^2
  + \lambda_{\text{KL}}\, D_{\text{KL}}\big( P \,\|\, Q(w) \big) .
\label{eq:joint_loss}
\end{equation}
Because the two terms are normalized differently, the effective ratio of KL to MSE is
$10^3$--$10^4$ at $\lambda_{\text{KL}}=0.1$: what is optimized is attention-map distillation with an
output-MSE regularizer (Appendix~\ref{app:lambda}).

\subsection{Why one joint objective rather than three}
\label{sec:joint}
Attention is \emph{bilinear} in $(W_Q, W_K)$: the scores depend on the pair only through
$W_Q W_K^\top$, whose perturbation contains a cross term $\delta_Q \delta_K^\top$ that no per-matrix
model represents. Treating the three separately is the block-diagonal model $H_{\text{sep}}$ of
$H_{\mathcal{L}}=\nabla^2_w\Ljab$, discarding cross-blocks $\Gamma = H_{\mathcal{L}} -
H_{\text{sep}}$ that are structurally non-zero: $\Ljab$ is invariant under $W_Q \mapsto W_Q R$,
$W_K \mapsto W_K R^{-\top}$ per head, so $H_{\mathcal{L}}$ has a null space of dimension at least
$n_{\text{head}} d_{\text{head}}^2$ ($49{,}152$ per GPT-2 block) on which the separable model
reports damage where there is none. With a zero-mean rounding residual of variance $\sigma^2$, the
two agree in expectation but not in spread:
\begin{equation}
\mathbb{E}[\Delta\Ljab] = \mathbb{E}[\Delta\Ljab^{\text{sep}}], \qquad
\operatorname{Var}[\Delta\Ljab] - \operatorname{Var}[\Delta\Ljab^{\text{sep}}]
  = \tfrac{1}{2}\sigma^4 \|\Gamma\|_F^2 \;>\; 0 .
\label{eq:var}
\end{equation}
A per-matrix model is therefore unbiased on average but over-confident about the tail --- the wrong
profile for allocation, since perplexity is convex and effectively unbounded in the low-bit
direction. The argument bears on reconstruction and fine-tuning; for scoring, a trace sees only
diagonal blocks and the two coincide (Appendix~\ref{app:joint}).

\subsection{Sensitivity and allocation}
\label{sec:jab}\label{sec:alloc}
The criterion is the Hessian trace of the \emph{same} loss w.r.t.\ the \emph{same} vector,
estimated with Hutchinson's method~\citep{hutchinson1990stochastic},
$\operatorname{Tr}(H_{\mathcal{L}}) \approx \frac1n\sum_i v_i^\top H_{\mathcal{L}} v_i$ with
Rademacher $v_i$, each product computed by two chained backward passes~\citep{pearlmutter1994hessian}.
Because $\Ljab$ is zero at the float point $w^\ast$, the Hessian there reduces to
\begin{equation}
H_{\mathcal{L}}(w^\ast) = 2 J_A^\top J_A + \lambda_{\text{KL}}\, J_Q^\top \mathcal{I}(P)\, J_Q ,
\label{eq:gn}
\end{equation}
a squared attention-output Jacobian plus an attention-map Fisher information, not a generic
curvature probe. Following HAWQ-V2, the allocator scores each candidate width by
\begin{equation}
\textbf{(C1)}\;\; \Omega_i(b) = \operatorname{Tr}(H_i)\,\big\|Q_b(W_i)-W_i\big\|_F^2 ,
\qquad
\textbf{(C2)}\;\; \Omega^{\text{prop}}_i(b) = (L-i)\,\Omega_i(b),
\label{eq:omega}
\end{equation}
where C2 re-weights by the number of layers an error propagates through. As a bound, the oracle
\textbf{(C3)} measures each $(i,b)$ pair's end-to-end logit KL by quantizing only layer $i$
($\approx73$ model loads). Each layer then chooses $b_i \in \mathcal{B}=\{2,3,4,8,16\}$ by
\begin{equation}
\min \textstyle\sum_i \Omega_i(b_i) \quad \text{s.t.} \quad \sum_i c(b_i)\le B ,
\label{eq:mckp}
\end{equation}
solved greedily or on the pooled Pareto frontier (Appendix~\ref{app:alloc}).

\subsection{Joint fine-tuning through the quantizer}
\label{sec:ste}
GPTQ solves~\eqref{eq:local_loss}, not~\eqref{eq:joint_loss}. The final stage takes GPTQ's solution
as a warm start and descends $\Ljab$ through the quantizer, using the straight-through estimator on
weights~\citep{bengio2013ste} and the LSQ gradient on the per-(row, group)
scales~\citep{esser2020lsq}. Because GPTQ's output sits exactly on grid points, learning rates are
set in units of the mean grid step $\overline{s}$ ($\eta_w = \alpha_w\overline{s}$,
$\eta_s = \alpha_s\overline{s}$) over 200 Adam steps; teacher and student see the same
partially-quantized input; a held-out batch gates every write-back; and we log the \emph{flip rate},
the fraction of weights whose grid position changed, since a stage that changes nothing and one that
changes everything can leave perplexity equally unmoved (Appendix~\ref{app:ste}).

\subsection{Metrics}
\label{sec:metrics}
We report perplexity and next-token accuracy under a sliding window, the weight-space error
$\varepsilon^W = \|W - \widehat{W}\|_F/\|W\|_F$, the attention-space error $\varepsilon^A$, and the
amplification factor $\rho = \overline{\varepsilon^{A}}/\overline{\varepsilon^{W}}$, which separates
weight fidelity from functional fidelity (Appendix~\ref{app:metrics}).

\section{Experimental Setup}
\label{sec:setup}

We study GPT-2 small~\citep{radford2019gpt2} (124M parameters, 12 blocks) and
Mistral-7B-v0.1~\citep{jiang2023mistral} (32 layers, grouped-query attention, rotary embeddings). Both use group size 128, activation ordering, $\lambda_{\text{KL}}=0.1$ unless ablated (except the MLP study, where it is 0 by construction), and 10 Hutchinson
probes per layer. The \textbf{attention-only} study quantizes the QKV projection: on GPT-2 it sweeps four criteria over seven budgets with and without fine-tuning, calibrating on WikiText-2~\citep{merity2017pointer} or C4~\citep{raffel2020c4}; on Mistral-7B it covers all four calibrate$\to$evaluate pairings. The \textbf{MLP-only} study quantizes GPT-2's \texttt{c\_fc} and \texttt{mlp.c\_proj}. The \textbf{full-model} study quantizes all seven linear modules of Mistral-7B. We evaluate GPT-2 on the full WikiText-2 test split and Mistral-7B on 32 windows of 512 tokens, so Mistral perplexities are exact within a study but not comparable to published full-split figures. All results are single-seed. Complete configurations are in Appendix~\ref{app:setup}.

\section{Attention-Only Quantization}
\label{sec:m-qkv}

\begin{table}[t]
\centering
\caption{Attention-only Mistral-7B at 3 bits, all four calibrate$\to$evaluate pairings
(W${=}$WikiText-2), with the share of the uniform-versus-fp16 gap closed. Full sweep:
Appendix~\ref{sec:allocres}.}
\label{tab:qkvbrief}
\small
\setlength{\tabcolsep}{6pt}
\begin{tabular}{@{}l ccc c c@{}}
\toprule
Pairing & fp16 & Uniform & Adaptive & Gap closed & $\Delta$ Acc. \\
\midrule
W $\to$ W    & 4.818 & 8.738  & \textbf{5.228} & $89.5\%$ & $+0.031$ \\
C4 $\to$ C4  & 7.492 & 13.506 & \textbf{8.898} & $76.6\%$ & $+0.014$ \\
W $\to$ C4   & 7.492 & 13.575 & \textbf{8.622} & $81.4\%$ & $+0.018$ \\
C4 $\to$ W   & 4.818 & 8.875  & \textbf{5.428} & $85.0\%$ & $+0.028$ \\
\bottomrule
\end{tabular}
\end{table}

\paragraph{At 7B scale, allocation carries the result}
On Mistral-7B, adaptive allocation dominates uniform GPTQ at 3 bits in all four corpus pairings, recovering $77$--$90\%$ of the uniform-versus-fp16 gap and $2$--$3$ accuracy points (Table~\ref{tab:qkvbrief}). By 4 bits every method is within $1$ -$2\%$ of fp16, and refinement is second-order: at 3 bits allocation is worth $3$--$5$ perplexity points and fine-tuning at most $0.17$ in either direction. The allocation transfers across corpus mismatch, and the GPT-2 trace ranking is nearly calibration-invariant (Appendix~\ref{sec:cross}).

\paragraph{A cheap criterion matches an expensive oracle}
Across the four GPT-2 budgets $\geq3.5$ bits, the Hutchinson criterion and the $\approx73$-load oracle differ by $0.17$ perplexity on average (Table~\ref{tab:sweep}). On Mistral, second-order scoring beats a first-order Fisher proxy at 3 bits ($5.268$ vs.\ $5.302$), while greedy and Pareto solvers are indistinguishable: the score matters, the solver does not.

\paragraph{Mixed precision is not uniformly safe, and the objective says why}
On GPT-2 at exactly $3.0$ bits, adaptive allocation reaches $86.36$ against uniform's $38.35$. This follows from the objective. At an average of exactly 3 bits, an assignment with no block below 3 must place every block at 3, which is the uniform solution; the adaptive assignment differs from uniform, so at least one block sits at 2 bits, and one is enough to dominate the total (uniform 2-bit scores $5137.6$). Minimizing a \emph{sum} of $\Omega_i(b_i)$, the allocator cannot see that perplexity is convex and unbounded below. Depth
re-weighting C2 halves the damage ($43.93$) at no cost. Mixed precision earns its keep at fractional
budgets instead, beating a geometric interpolation between neighboring uniform solutions by
$2.2\times$ at 2.5 bits and $1.11\times$ at 3.5 (Appendix~\ref{sec:allocres}).

\paragraph{Refinement recovers attention behavior, not weights}
On GPT-2, joint fine-tuning improves 18 of 21 configurations, and does so by moving the weights
\emph{away} from float: at 4 bits $\varepsilon^W$ rises $19\%$ ($0.174\to0.208$) while
$\varepsilon^A$ falls $33\%$ ($0.238\to0.160$), and both perplexity and accuracy improve. Every
GPTQ-only arm has $\rho>1$ and every fine-tuned arm $\rho<1$, with the ordering
$\rho_{\text{GPTQ}} > \rho_{\text{MSE-only}} > \rho_{\text{MSE+KL}}$ holding at every budget on both
corpora (Appendix~\ref{sec:errspace}). The attention \emph{map} is what must be preserved: the KL
term supplies $88\%$ of the gain, and MSE-only refinement lowers $\varepsilon^A$ yet still degrades
perplexity.

\section{Beyond Attention: the MLP}
\label{sec:m-mlp}

The JAB machinery is not intrinsically tied to the softmax. To test whether it generalizes beyond
attention, we apply the same allocation-and-refinement pipeline to the GPT-2 MLP, quantizing
\texttt{mlp.c\_fc} and \texttt{mlp.c\_proj} with attention kept in fp32; the MLP holds
$66.6\%$ of the non-embedding weights against $25.0\%$ for QKV. Two structural differences follow.
The two matrices are sequential rather than a jointly coupled projection, so their Hessians are
collected in execution order; and there is no attention map to distill, so the KL component is
absent ($\lambda_{\mathrm{KL}}=0$) and $\Ljab$ reduces to output MSE. The experiment therefore tests
which parts of JAB transfer once its attention-specific signal is removed.

\begin{table}[t]
\centering
\caption{GPT-2 MLP, WikiText-2 perplexity (fp32 control $24.357$) at the two budgets with a uniform
baseline. Full sweep: Appendix~\ref{sec:mlp}, Table~\ref{tab:mlp}.}
\label{tab:mlpbrief}
\small
\setlength{\tabcolsep}{8pt}
\begin{tabular}{@{}c ccc@{}}
\toprule
$B$ (bits) & Uniform & JAB (C1) & Oracle (C3) \\
\midrule
3.0 & 53.71 & 120.70 & \textbf{46.15} \\
4.0 & \textbf{26.23} & 27.46 & \textbf{26.23} \\
\bottomrule
\end{tabular}
\end{table}

\paragraph{The Hessian criterion does not improve over uniform allocation, whereas the oracle does}
At the two budgets for which a uniform baseline is available, the Hessian criterion performs worse
than uniform quantization: $120.70$ vs.\ $53.71$ perplexity at 3 bits and $27.46$ vs.\ $26.23$ at
4 bits (Table~\ref{tab:mlpbrief}). This failure differs qualitatively from the QKV collapse, since the 4-bit allocation contains no unit below 3 bits. The oracle, in contrast, improves on uniform at 3 bits, reaching $46.15$, a $14\%$ reduction. Thus, useful allocation structure exists in the MLP, but the Hessian criterion fails to identify it. 

\paragraph{The oracle allocates by role, not depth}
At every fractional budget the oracle returns the same \emph{flat} allocation in every block, split only by matrix role: \texttt{mlp.c\_fc} at $b{+}1$ and \texttt{mlp.c\_proj} at $b{-}1$ (Appendix~\ref{sec:mlp}, Table~\ref{tab:mlpalloc}). The criterion cannot find that structure, by construction. One trace over the concatenated block vector is identical for
both units of a block, so the only thing that can tell them apart is the perturbation factor of Eq.~(\ref{eq:omega}). What the criterion sees instead is a depth gradient rising $4.5\times$, unlike the QKV trace on the same network, which peaks at block 3 with a $41\times$ spread. Sensitivity profiles are a property of the sublayer, not of the architecture alone.
\section{Full-Model Quantization}
\label{sec:m-full}

We now put QKV and MLP under one shared budget, first on GPT-2 small, then on every linear module of Mistral-7B.

\subsection{GPT-2: QKV and MLP Together}
\label{sec:g-full}
The fused QKV projection and both MLP matrices of every GPT-2 block now share one budget: 36 units, $91.7\%$ of block weights, giving $2.21\times$ end-to-end compression at 4 bits against $1.66\times$ for the MLP alone. Four allocators are compared at five budgets, calibrated on C4, GPTQ-only, evaluated on WikiText-2 (fp32 control $24.357$) and a disjoint C4 slice with the same ranking (Appendix~\ref{sec:gpt2c4}). Sensitivity tables are normalized per block type first, since raw JAB scores are not commensurate across types. The shared block trace was also replaced by an unbiased per-unit trace in one of the sweeps, but no improvement was obtained.

\begin{table}[t]
\centering
\caption{GPT-2 with QKV and MLP quantized together (C4 calibration, GPTQ-only): WikiText-2
perplexity (fp32 control $24.357$) and accuracy at the two budgets that separate the allocators.
Full table: Appendix~\ref{sec:gpt2wiki}; in-domain C4, same ranking: Appendix~\ref{sec:gpt2c4}.}
\label{tab:gpt2brief}
\small
\setlength{\tabcolsep}{8pt}
\begin{tabular}{@{}l cc cc@{}}
\toprule
 & \multicolumn{2}{c}{$B=3.5$} & \multicolumn{2}{c}{$B=4.5$} \\
\cmidrule(lr){2-3}\cmidrule(l){4-5}
Allocator & PPL & Acc & PPL & Acc \\
\midrule
JAB (C1)          & 75.009 & 0.296 & 29.041 & 0.394 \\
JAB${+}$prop (C2) & 67.337 & 0.307 & 28.694 & \textbf{0.397} \\
Type-offset       & \textbf{64.896} & \textbf{0.310} & \textbf{28.211} & 0.395 \\
\bottomrule
\end{tabular}
\end{table}

\paragraph{Role beats curvature once both sublayers share the budget}
Type-offset needs no scoring pass at all and still leads at the fractional budgets: $64.896$ at $3.5$ bits and $28.211$ at $4.5$, against JAB${+}$prop's $67.337$ and $28.694$ (Table~\ref{tab:gpt2brief}). JAB's attention-defined sensitivity over-invests in QKV and under-funds the MLP, which now dominates the parameter count; only near saturation, at 6 bits, does uniform win. Two observations carry over to Mistral (Appendix~\ref{sec:gpt2wiki}): $\varepsilon^A$ stops tracking perplexity once the MLP shares the budget, and the joint STE stage raises perplexity at every budget, by $16.35\times$ at 3 bits.

\subsection{Mistral-7B: All Seven Modules Together}
\label{sec:m-full-mistral}
We now quantize every linear module of all 32 Mistral-7B layers under one budget: 224 units, $6.98$B parameters, $96.4\%$ of the model.

\paragraph{A parameter-weighted budget}
The sublayer studies had units of equal size, so per-unit and per-parameter costs coincided. Here grouped-query attention makes \texttt{k\_proj} and \texttt{v\_proj} ($4.2$M) a quarter of \texttt{q\_proj} and \texttt{o\_proj} ($16.8$M), while the three SwiGLU matrices hold $58.7$M each --- a $14\times$ span, with the MLP carrying $80.8\%$ of the weights. A per-unit cost would price those equally, so we define the budget per parameter,
\begin{equation}
  B \;=\; \frac{\sum_i P_i\, b_i}{\sum_i P_i}, \qquad c(b_i) = P_i\, b_i ,
  \label{eq:pwbudget}
\end{equation}
with $P_i$ the parameter count of unit $i$. On the quantized weights this makes storage $16/B$ times smaller than fp16: $13.96$\,GB $\to 3.93$\,GB ($3.56\times$) at $B=4.5$.

\paragraph{Type-offset: a Hessian-free role prior}
Motivated by the MLP oracle's role rule, Type-offset gives each unit kind $\kappa$ a fixed integer offset $\Delta_\kappa$ and assigns
\begin{equation}
  b_i \;=\; \mathrm{nearest}_{\mathcal{B}}\!\left(t^\star + \Delta_{\kappa(i)}\right),
  \qquad t^\star = \arg\min_t \Bigl|\,\textstyle\sum_i c(b_i) - B \sum_i P_i\Bigr| ,
  \label{eq:typeoffset}
\end{equation}
followed by single-unit moves meeting the budget exactly: seven integers, no scoring pass. We set $\Delta = 0$ for $q,k,v,o$, $+1$ for \texttt{gate} and \texttt{up}, $-1$ for \texttt{down}, transposed from the GPT-2 oracle rather than tuned, and also test \emph{JAB-norm}, which divides each trace by $\overline{\mathrm{diag}(H_i)}$.

\begin{table}[t]
\centering
\caption{Mistral-7B, all seven modules quantized: WikiText-2 perplexity (fp16 control $6.643$) and
compression of the quantized weights. Budgets $B$ are bits per parameter (Eq.~(\ref{eq:pwbudget})).
\textbf{Left:} criteria at fractional budgets; Type-offset needs no scoring pass. \textbf{Right:}
uniform quantization, defined only at integer budgets, and the adaptive JAB arm. Best per budget in bold.}
\label{tab:fullmodel}
\small
\begin{minipage}[t]{0.55\linewidth}
\centering
\setlength{\tabcolsep}{5pt}
\begin{tabular}[t]{@{}l ccc@{}}
\toprule
$B$ (bits/param) & 2.5 & 3.5 & 4.5 \\
Compression & $6.40\times$ & $4.57\times$ & $3.56\times$ \\
\midrule
Type-offset & 6624.3 & 32.33 & \textbf{6.970} \\
JAB (C1)    & 1799.6 & \textbf{8.570} & 7.250 \\
JAB-norm    & \textbf{648.0} & 109.6 & 10.77 \\
\bottomrule
\end{tabular}
\end{minipage}\hfill
\begin{minipage}[t]{0.42\linewidth}
\centering
\setlength{\tabcolsep}{5pt}
\begin{tabular}[t]{@{}l c c r@{}}
\toprule
Arm & $B$ & Compr. & PPL \\
\midrule
Uniform      & 2.0 & $8.00\times$ & 64779.9 \\
Uniform      & 3.0 & $5.33\times$ & 9.126 \\
Uniform      & 4.0 & $4.00\times$ & 6.998 \\
JAB adaptive & 4.3 & $3.72\times$ & 7.373 \\
\bottomrule
\end{tabular}
\end{minipage}
\end{table}

\subsection{Layer Role Outperforms Curvature}
\label{sec:m-role}
\paragraph{The role prior wins at a matched budget}
At $B=4.5$ bits per parameter, Type-offset reaches $6.970$ perplexity against $7.250$ for JAB and $10.770$ for JAB-norm (Table~\ref{tab:fullmodel}). All three realize the same budget to three decimals, $4.500$ bits per parameter, but Type-offset computes no Hessian and runs no Hutchinson probes. The curvature information JAB pays for does not buy a better allocation than seven integers. The MLP experiment showed the same thing: there too the end-to-end oracle preferred a flat allocation set by matrix role rather than depth. That it recurs in Mistral-7B, with grouped-query attention, SwiGLU and $58\times$ the parameters, is what makes us read it as a property of heterogeneous transformers rather than of one architecture.

\paragraph{Why the JAB criterion misses the role structure}
JAB computes one Hessian trace for the concatenated parameter vector of each decoder block, so all seven linear modules in a block receive the same trace value. The criterion can separate them only through the perturbation term in Eq.~(\ref{eq:omega}). What it does vary over is depth: the trace rises $381\times$ from block 0 to block 31. The oracle's allocation has no depth structure at all, only a role split, so JAB spends bits along the wrong axis. It distinguishes \emph{where} a layer sits rather than lemph{which} matrix it is. A per-unit trace would remove this limitation (Appendix~\ref{sec:fullmodel}).
Section~\ref{sec:m-floor} removes this limitation and finds the ranking unchanged.

\paragraph{Type-offset is useful, but its offsets are not universally transferable}
Type-offset degrades sharply below $4.5$ bits: at $B=3.5$ the fixed $\Delta_{\mathrm{down}}=-1$ offset assigns 2 bits to $30$ of the $32$ $\mathrm{down}_{\mathrm{proj}}$ matrices, giving $32.33$ perplexity. The GPT-2 MLP experiment pointed to this direction, but Mistral's $\mathrm{down}_{\mathrm{proj}}$ receives heavier-tailed inputs, so the same offset need not apply; the floor of Section~\ref{sec:m-floor} removes the failure entirely. Each candidate offset costs one evaluation, so we report the $4.5$-bit result as a floor for the role-prior family, not a tuned choice.

\subsection{Block-Local Objectives Can Diverge from End-to-End Quality}
\label{sec:m-proxygap}

\begin{table}[t]
\centering
\caption{Joint refinement of the adaptive $B{=}4.3$ allocation. Block~0 is the only block whose
weights change, so refining blocks 0--3 reproduces the full-depth result exactly. Full table:
Appendix~\ref{sec:fullmodel}.}
\label{tab:ftbrief}
\small
\setlength{\tabcolsep}{7pt}
\begin{tabular}{@{}l c c r@{}}
\toprule
Refinement & Blocks refined & Blocks changed & PPL \\
\midrule
None (GPTQ only)  & --    & --        & \textbf{7.373} \\
Joint FT          & 0--31 & 1 (block 0) & 239.52 \\
Joint FT          & 0--3  & 1 (block 0) & 239.52 \\
\bottomrule
\end{tabular}
\end{table}

\paragraph{A locally improved block can substantially degrade the complete model}
Joint refinement of the adaptive $B=4.3$ allocation reduces the block-local objective of block~0 from $1.85$ to $0.41\times10^{-3}$, a $4.6\times$ improvement. End-to-end perplexity increases from $7.373$ for GPTQ-only to $239.52$ after refinement. Refining only blocks 0--3 gives exactly the same perplexity and the same error metrics (Table~\ref{tab:ftbrief}), and in both runs block~0 is the only block whose quantized weights move at all; the other 31 return to their GPTQ grid points.

This is not a failure of the local optimization. The refinement reduces the objective it was given and passes the held-out checkpoint gate. One refined block still raises perplexity $32.5\times$.



\paragraph{Why block 0 is especially vulnerable}
Two properties coincide in this block. The adaptive allocator assigns 2 bits to six of its seven linear modules, giving a weight-space error of $\varepsilon^W=1.015$: the quantized block is already farther from its floating-point weights than the zero matrix is, in relative Frobenius error. And the output of block~0 passes through all 31 downstream decoder blocks. Refinement is therefore working on a severely quantized representation at the point of maximum downstream leverage.

We read the failure as follows. Minimizing a block-local objective on a small calibration set can move the weights toward calibration-specific behavior while losing what the downstream network needs. The held-out gate cannot catch this, because it evaluates the same local objective rather than the end-to-end metric.

Granularity matters too. Under uniform 4-bit quantization the same refinement stage changes three blocks and raises perplexity by $1.6\times$ rather than $32.5\times$, so in these two runs the mismatch is far more severe when low-bit assignments are concentrated.


\paragraph{Implication for refinement}
The problem is not specific to $\Ljab$. Any refinement procedure that optimizes a block-local target and validates against that same target gives no guarantee of better end-to-end quality. At minimum, refinement should track a global metric alongside a per-block flip-rate diagnostic (Appendix~\ref{app:ste}, Eq.~(\ref{eq:flip})), which is what identifies a locally accepted update with outsized downstream effect. The same block also generalizes the 2-bit pathology we prove for GPT-2 in Section~\ref{sec:m-qkv}. The additive objective~\eqref{eq:mckp} cannot see it, and it appears whenever the allocator is free to concentrate low widths on a single block.

\subsection{A 3-Bit Floor with Per-Unit Curvature}
\label{sec:m-floor}
Each diagnostic suggests a change: the additive objective cannot see the low-bit cliff, so we impose
the floor $b_i \ge 3$; and the shared block trace cannot rank units within a block, so we replace it
with an unbiased \emph{per-unit} trace at no extra cost, partitioning $v^\top\!Hv$ over each unit's
coordinate range (Appendix~\ref{sec:perunit}).

\begin{table}[tb]
\centering
\caption{Mistral-7B, all seven modules, floor $b_i\ge3$ and per-unit traces (GPTQ-only; fp16
$6.643$; uniform 4-bit $6.938$; adaptive $B{=}4.3$ $7.211$). Compression is on the quantized
weights. Not comparable to Table~\ref{tab:fullmodel} (64 vs 128 Hessian batches);
see Appendix~\ref{sec:perunit}.}
\label{tab:floor}
\small
\setlength{\tabcolsep}{7pt}
\begin{tabular}{@{}c c ccc@{}}
\toprule
$B$ & Compr. & JAB (C1) & JAB-norm & Type-offset \\
\midrule
3.0 & $5.33\times$ & 9.713 & 9.713 & 9.713 \\
3.5 & $4.57\times$ & \textbf{7.514} & 8.215 & 7.875 \\
4.5 & $3.56\times$ & 7.158 & 7.592 & \textbf{6.933} \\
\bottomrule
\end{tabular}
\end{table}

\paragraph{The floor removes the catastrophic regime}
No arm now exceeds $9.72$, against $32.33$ and $109.6$ at $B=3.5$ without it; at $B=3.0$ the floor
saturates the budget, so all four criteria return the uniform assignment and the identical $9.713$.
The usable operating points are Type-offset at $B=3.5$ ($7.875$, $4.57\times$) and $B=4.5$ ($6.933$,
$3.56\times$): $14$\,GB reduced to $3.93$\,GB for $4.4\%$ higher perplexity than fp16.

\paragraph{Per-unit curvature does not change the ranking}
The traces now discriminate strongly within a block, yet Type-offset still leads at matched $B=4.5$,
$6.933$ against $7.158$: $q$ and $k$ take $77\%$ of the trace mass while holding $19.2\%$ of the
weights, since $\Ljab$ is defined on the attention output and map. The limitation is the objective
the curvature comes from, not the shared trace this experiment removes (Appendix~\ref{sec:perunit}).

\section{Conclusion}
\label{sec:conclusion}


Using one loss for both reconstruction and allocation lets us analyze the two stages of
mixed-precision PTQ in a single frame, and on attention it works: adaptive allocation recovers most of the uniform-to-fp16 gap at 7B. Beyond attention it does not transfer. Where units are heterogeneous, matrix role carries more signal than the curvature we computed, and per-unit resolution does not change that, since the curvature comes from an attention objective. Two lessons hold independently of JAB. An additive objective cannot see the low-bit cliff, and a floor $b_i\ge3$ repairs it at no cost. And a certified local improvement is not evidence of a global one (Appendix~\ref{sec:limits}).

Both regimes matter, but not equally. Compression is the reason to quantize, and the ratio moves fastest at the low end: $16/3$ is $5.3\times$ against $3.56\times$ at $4.5$ bits. Large models are the ones forced to live there, and $3.5$ bits is where the curvature criterion is ahead of the role prior ($7.51$ against $7.87$). The regime JAB serves is therefore the one that grows in importance with model size, although we test only to 7B.
\subsection*{AI use statement}
In this work, we used generative AI tools (a large language model assistant) for editing
and condensing the manuscript text, drafting and checking LaTeX, writing experiment, analysis
and plotting code, and cross-checking the numbers reported in the text against our result
files. We have not used generative AI tools to produce any experimental result: every number
in the paper is computed by our code from the models' outputs, and every AI-drafted script
was run and checked by the authors.
We have reviewed all AI-assisted work and take responsibility for the final content of this
work, including text, claims or artifacts produced with the aid of generative AI.
\section*{Reproducibility Statement}
Code, configurations and seeds: \href{https://anonymous.4open.science/r/Attention-Aware-Joint-Mixed-Precision-Quantization-of-Transformer-Models-6080}{anonymous.4open.science/r/\dots-6080}.
All runs use seed 42 and are single-seed; full hyperparameters are in Appendix~\ref{app:setup}.

\bibliography{references}
\bibliographystyle{iclr2027_conference}
\clearpage  
\appendix
\section{Positioning Against Prior Work}
\label{app:related}

Table~\ref{tab:comparison} contrasts the reconstruction objective each method minimizes with the importance criterion it allocates by. JAB is the only entry for which the two are the same functional; the oracle bounds what any criterion can achieve, and Type-offset (Section~\ref{sec:m-full}) sidesteps the question by using no criterion at all.

\begin{table}[t]
\centering
\caption{Relationship between reconstruction objective and importance
criterion. JAB closes the gap for attention; the type-offset baseline
sidesteps it entirely.}
\label{tab:comparison}
\scriptsize
\setlength{\tabcolsep}{1.5pt}
\begin{tabular}{
    >{\raggedright\arraybackslash}p{0.17\columnwidth}
    >{\raggedright\arraybackslash}p{0.28\columnwidth}
    >{\raggedright\arraybackslash}p{0.29\columnwidth}
    c}
\toprule
Method & Recon.\ objective & Importance criterion & Same? \\
\midrule
GPTQ~\citep{frantar2023gptq} & $\|XW - X\widehat{W}\|^2$/layer & N/A (uniform bits) & -- \\
HAWQ-V2~\citep{dong2020hawqv2} & $\|XW - X\widehat{W}\|^2$/layer & $\operatorname{Tr}(H)$ of task loss & No \\
APTQ~\citep{guan2024aptq} & Weight recon.\ ${+}$ attn.\ grad. & Hessian trace, weight-space & Related \\
JAB-Hessian (ours) & Joint QKV attn.\ loss $\Ljab$ & $\operatorname{Tr}(\nabla^2 \Ljab)$ & Identical \\
Fisher-coupling (ours) & Joint QKV attn.\ loss $\Ljab$ & $\|\nabla \Ljab\|^2$ at 2-bit probe & First-order \\
Oracle (ours) & Joint QKV attn.\ loss $\Ljab$ & End-to-end logit KL & Bound \\
Type-offset (ours) & Layer-local GPTQ & Role-based offset table & Hessian-free \\
\bottomrule
\end{tabular}
\end{table}

\section{Method Details}
\label{app:method}

\subsection{GPTQ core}
\label{app:gptq}
\paragraph{OBS update}
OBS gives the closed form for rounding weight $w_q$ while optimally compensating the not-yet-quantized weights $F$:
\begin{equation}
\delta_F = - \frac{w_q - Q(w_q)}{[H^{-1}]_{qq}} \, H^{-1}_{F,q},
\label{eq:obs}
\end{equation}
i.e.\ for column $i$, $\text{err} = (w_{\text{col}} - q_{\text{col}})/H^{-1}[i,i]$ and $W[:, i{+}1{:}] \mathrel{-}= \operatorname{outer}(\text{err}, H^{-1}[i, i{+}1{:}])$. Both standard refinements are used: group-wise scaling and activation ordering ($\operatorname{diag}(H)$-descending column permutation).

\paragraph{Numerical precision is load-bearing}
The column loop divides by an inverse-Hessian diagonal $d_{\text{in}}$ times per row, so fp32 rounding compounds and produces \emph{non-monotone} perplexity across bit-widths. GPT-2 can afford a float64 solve; Mistral cannot (float64 is $32\times$ slower on a T4 over a $4096$-step loop), so the streaming build accumulates $H$ in float64 but solves in float32 --- a caveat carried into Section~\ref{sec:limits}.

\subsection{Effective weight of the KL term}
\label{app:lambda}
\paragraph{The nominal $\lambda_{\text{KL}}$ is not the effective weight}
$\mathcal{L}_{\text{MSE}}$ takes an element-wise mean over $B\!\cdot\!T\!\cdot\!d$ entries, whereas $\mathcal{L}_{\text{KL}}$ uses \texttt{batchmean}, dividing only by $B$ while summing over $B\!\cdot\!n_{\text{head}}\!\cdot\!T^2$. The measured ratio $\lambda_{\text{KL}}\mathcal{L}_{\text{KL}}/\mathcal{L}_{\text{MSE}}$ is $10^3$--$10^4$ at $\lambda_{\text{KL}}=0.1$, so what is optimized is \emph{attention-map distillation with an output-MSE regularizer}. We keep the implemented convention and quantify it by ablation (Section~\ref{sec:ft}), since it carries most of the benefit.

\subsection{Why one joint objective rather than three: full argument}
\label{app:joint}
Section~\ref{sec:joint} states the result; this subsection gives the argument.

\paragraph{It cannot be worse}
The grid is a product set $\mathcal{G} = \mathcal{G}_Q \times \mathcal{G}_K \times \mathcal{G}_V$ --- each matrix rounds on its own scales, and no choice for one constrains the others --- so any separately obtained triple is feasible for the joint problem and
\begin{equation}
\min_{w \in \mathcal{G}} \Ljab(w) \;\le\; \Ljab\big(\widehat{W}^{\text{sep}}_Q, \widehat{W}^{\text{sep}}_K, \widehat{W}^{\text{sep}}_V\big).
\label{eq:feas}
\end{equation}
The content lies in when \eqref{eq:feas} is strict --- and in the fact that a separate pipeline does not minimize $\Ljab$ at all, but a sum of per-matrix surrogates, so two gaps stack.

\paragraph{What separability discards}
Partition $H_{\mathcal{L}}$ into $3\times3$ blocks indexed by $Q,K,V$. Separate treatment is exactly the block-diagonal model $H_{\text{sep}} = \operatorname{blockdiag}(H_{QQ}, H_{KK}, H_{VV})$, and the error it commits at second order is exactly the cross-blocks,
\begin{equation}
\Delta\Ljab - \Delta\Ljab^{\text{sep}} = \delta_Q^\top H_{QK}\delta_K + \delta_Q^\top H_{QV}\delta_V + \delta_K^\top H_{KV}\delta_V .
\label{eq:cross}
\end{equation}
These do not vanish, because attention is \emph{bilinear} in $(W_Q, W_K)$: the scores depend on that pair only through $M = W_Q W_K^\top$, and $\widehat{W}_Q \widehat{W}_K^\top = M + \delta_Q W_K^\top + W_Q \delta_K^\top + \delta_Q \delta_K^\top$, whose last term no per-matrix model can represent. In the Gauss--Newton form \eqref{eq:gn}, $H_{QK} = 2 J_A^{Q\top} J_A^{K} + \lambda_{\text{KL}} J_Q^{Q\top}\mathcal{I}(P) J_Q^{K}$ --- a Gram cross-block that vanishes only if perturbing $W_Q$ and $W_K$ moved the attention output in orthogonal directions. They do not: both move it through $M$.

\paragraph{The discarded term is unbounded in relative terms}
Take one head with $d_{\text{head}} = 1$, so $W_Q, W_K$ reduce to vectors $q, k$ and the block depends on them only through $qk^\top$; measure damage by $\|\Delta(qk^\top)\|_F^2$. With $\delta_q = \varepsilon q$ and $\delta_k = -\tfrac{\varepsilon}{1+\varepsilon}k$ the product is reproduced \emph{exactly}: true damage $0$, separable estimate $\Theta(\varepsilon^2)\|qk^\top\|_F^2$ --- an infinite relative error at any $\varepsilon$. With $\delta_q = \varepsilon q$, $\delta_k = \varepsilon k$ the true damage is $\approx 4\varepsilon^2\|qk^\top\|_F^2$ against a separable estimate of $2\varepsilon^2\|qk^\top\|_F^2$ --- an underestimate by exactly $2\times$. Identical per-matrix perturbation norms; true damage anywhere from zero to twice the separable prediction, decided entirely by relative alignment, which is what a per-matrix model cannot see.

This is not pathological: $\Ljab$ is exactly invariant under $W_Q \mapsto W_Q R$, $W_K \mapsto W_K R^{-\top}$ per head, so $H_{\mathcal{L}}$ is singular with a null space of dimension at least $n_{\text{head}} d_{\text{head}}^2 = 49{,}152$ per GPT-2 block. Every such direction has nonzero $Q$ \emph{and} $K$ components, so on it $\delta^\top H_{\mathcal{L}} \delta = 0$ while $\delta^\top H_{\text{sep}} \delta > 0$. Nor is the separable model conservative: if $H_{QK} \ne 0$, take $u,v$ with $u^\top H_{QK} v = c \ne 0$; then $(u, v, 0)$ and $(u, -v, 0)$ have identical block-diagonal forms while their true forms differ by $\pm 2c$, so it over- and under-estimates in equal measure.

\paragraph{Unbiased in the mean, over-confident in the tail}
Model the rounding residual as zero-mean and uncorrelated with variance $\sigma^2 = s^2/12$. The cross-blocks contribute no diagonal, so $\operatorname{Tr}(H_{\text{sep}}) = \operatorname{Tr}(H_{\mathcal{L}})$ and $\mathbb{E}[\Delta\Ljab] = \tfrac{\sigma^2}{2}\operatorname{Tr}(H_{\mathcal{L}}) = \mathbb{E}[\Delta\Ljab^{\text{sep}}]$: the two agree exactly \emph{in expectation}. Their variances do not. With $\Gamma = H_{\mathcal{L}} - H_{\text{sep}}$, the shared diagonal gives $\|H_{\mathcal{L}}\|_F^2 = \|H_{\text{sep}}\|_F^2 + \|\Gamma\|_F^2$, so
\begin{equation*}
\operatorname{Var}[\Delta\Ljab] - \operatorname{Var}[\Delta\Ljab^{\text{sep}}] = \tfrac{1}{2}\sigma^4 \|\Gamma\|_F^2 \;>\; 0 .
\end{equation*}
A per-matrix model is thus unbiased on average and systematically \emph{over-confident} about the tail, by exactly the squared norm of the cross-blocks --- the worst profile here, since Section~\ref{sec:allocres} shows perplexity is convex and effectively unbounded in the low-bit direction, so what an allocator must avoid is the tail, not the mean.

\paragraph{Scope of the argument}
Two limits matter. First, this bears on the reconstruction and fine-tuning half only: a trace sees only diagonal entries, so $\operatorname{Tr}(H_{\mathcal{L}}) = \operatorname{Tr}(H_{QQ}) + \operatorname{Tr}(H_{KK}) + \operatorname{Tr}(H_{VV})$ identically, and for the \emph{scoring} half joint and separate coincide exactly --- the criterion's advantage over HAWQ-V2 lies in which loss the diagonal blocks come from, not in jointness. Second, the argument concerns $\Ljab$, not perplexity, and $\|\Gamma\|_F/\|H_{\mathcal{L}}\|_F$ was never measured (L5). The decisive test is cheap, and \eqref{eq:var} predicts its signature: equal means, heavier tail.

\subsection{Fisher-coupling}
\label{app:fisher}
\paragraph{Fisher-coupling (first-order alternative)}
Since $\nabla_w\Ljab(w^\ast)=0$ identically, a gradient-based score must be evaluated \emph{off} the float point: Fisher-coupling scores layer $i$ by the squared gradient norm of $\Ljab$ at a coarse round-to-nearest 2-bit probe. It is cheaper than a double-backward HVP but is \emph{not} a decomposition of $H_{\mathcal{L}}$ --- a first-order proxy taken near the low-bit operating point. Section~\ref{sec:sens} compares the two.

\subsection{Allocators}
\label{app:alloc}
MCKP is NP-hard and exhaustive search ($5^{12}$ here, $5^{32}$ on Mistral) is intractable, so two heuristics are compared, both consuming the same $\Omega_i(b)$ table. \textbf{Greedy}: start every layer at the highest width, repeatedly apply the downgrade minimizing $\Delta\Omega/\Delta c$ until the budget is met, then spend leftovers on the best upgrade --- the classical fractional-knapsack walk \citep{dantzig1957discrete}, optimal only for the continuous relaxation. \textbf{MCKP/Pareto}: restrict each layer's $(b,\Omega)$ points to their Pareto-efficient frontier first, then walk down the pooled frontier by best ratio, matching HAWQ-V2's formulation more closely.

\subsection{Joint STE fine-tuning: the five design decisions}
\label{app:ste}
GPTQ solves \eqref{eq:local_loss}, not \eqref{eq:joint_loss}. The final stage takes GPTQ's solution as a warm start and descends $\Ljab$ \emph{through} the quantizer. Five decisions distinguish it from a naive STE loop.

\textbf{(1) STE on weights, LSQ on scales.} Two quantities need two gradient rules: the fake-quantized weight $\widehat{w}$ of \eqref{eq:quant}, and the per-group scale $s$ that defines the grid. The weight uses the straight-through estimator \citep{bengio2013ste} --- gradients pass through the rounding step unchanged --- while the scale is a learnable parameter that can widen or narrow the grid, updated with the LSQ gradient \citep{esser2020lsq}. With $v=w/s$ and $q=\operatorname{clip}(\operatorname{round}(v),\pm q_{\max})$,
\begin{equation}
\frac{\partial \widehat{w}}{\partial w} = 1, \qquad
\frac{\partial \widehat{w}}{\partial s} =
\begin{cases} q - v, & |v| \le q_{\max},\\ \operatorname{sign}(v)\,q_{\max}, & \text{else,}\end{cases}
\label{eq:lsq}
\end{equation}
the scale gradient further scaled by $(n_w q_{\max})^{-1/2}$ as in LSQ. One scale is kept per \emph{(output row, group)} rather than per weight: a per-weight scale would let $q_i s_i$ take any real value, silently restoring full precision and defeating the bit-width constraint.

\textbf{(2) Grid-relative learning rates.} GPTQ's output sits exactly on grid points, so a latent weight must travel $\ge s/2$ before $\operatorname{round}(\cdot)$ changes. Absolute rates are therefore meaningless; we set
\begin{equation}
\eta_w = \alpha_w \overline{s}, \qquad \eta_s = \alpha_s \overline{s},
\label{eq:lr}
\end{equation}
with $\overline{s}$ the mean grid step, $\alpha_w=0.05$, $\alpha_s=0.02$, cosine-annealed over 200 Adam \citep{kingma2015adam} steps.

\textbf{(3) Teacher and student share the same input.} Targets $A(X),P(X)$ use the block's \emph{float} weights on the \emph{same} $X$ the student sees, taken from the partially-quantized model --- GPTQ's own $\text{target}=W_{\text{float}}X_{\text{quant}}$ convention. A parallel all-float trajectory would fold upstream error, which block $i$ cannot fix, into its residual. \textbf{(4) Held-out checkpoint selection with a commit gate.} One calibration batch is held out of the training pool; every 10 steps the current weights are scored on it, the best checkpoint retained, and written back \emph{only} if it beats the GPTQ-only start. \textbf{(5) Flip-rate diagnostic.} Since a stage that changes nothing and one that changes everything can both leave perplexity unmoved, we log
\begin{equation}
\text{flip}_i = \tfrac{1}{d'}\big|\{ j : \operatorname{round}(w^{\text{GPTQ}}_j/s_j) \ne \operatorname{round}(w^{\text{FT}}_j/s_j) \}\big| .
\label{eq:flip}
\end{equation}

\paragraph{Why these decisions matter}
An earlier build used a fixed absolute rate and recorded a flip rate of $0.000$ on eleven of twelve blocks: each step moved weights less than a tenth of the way to the nearest grid boundary, so no quantized value could change regardless of gradient quality. The grid-relative rate of \eqref{eq:lr} over 200 steps raises the flip rate to $0.12$--$0.18$. When fine-tuning through a quantizer warm-started from a grid-exact solution, the learning rate must be expressed in units of the grid, and the flip rate must be logged.

\subsection{Evaluation metrics in full}
\label{app:metrics}
All four are computed in the same quantize-and-evaluate pass, after quantization and any fine-tuning.

\textbf{M1 --- Perplexity.} $\text{PPL} = \exp\big(\frac1N\sum_i \text{NLL}_i\big)$ under the sliding-window protocol, only the non-overlapping target span contributing, so no token is scored twice and every token carries $\ge$512 tokens of context. \textbf{M2 --- Next-token top-1 accuracy.} $\text{Acc} = \frac1N\sum_i \mathbb{1}[\arg\max_v p_\theta(v|x_{<i}) = x_i]$, on the identical windows, mask and forward pass as M1 --- a \emph{decision-level} metric, insensitive to how mass is spread over the rest of the vocabulary. \textbf{M3 --- Weight-space quantization error.} $\varepsilon^{W}_i = \|W_i - \widehat{W}_i\|_F / \|W_i\|_F$, block-averaged: pure weight space, no data, and up to normalization exactly what GPTQ's objective \eqref{eq:local_loss} and the perturbation term of \eqref{eq:omega} measure. \textbf{M4 --- Attention-space reconstruction error.} $\varepsilon^{A}_i$, the relative Frobenius error of the merged pre-\texttt{c\_proj} attention output, teacher and student evaluated on the \emph{same} $X$ from the live partially-quantized model, averaged over $|\mathcal{E}|=8$ fixed 512-token spans.

\textbf{Derived --- amplification factor.} $\rho = \overline{\varepsilon^{A}}/\overline{\varepsilon^{W}}$. M3 and M4 are relative errors of the same form on either side of the attention operator, so $\rho>1$ means causal softmax attention \emph{amplifies} the relative weight perturbation it is handed and $\rho<1$ that it \emph{attenuates} it. Reporting both makes the report's premise falsifiable: if attention space is the real objective and weight space only a proxy, then a pipeline optimizing the former should move \emph{away} from the float weights (M3 up) while moving \emph{toward} the float attention output (M4 down), and improve M1 and M2 while doing so.

\section{Experimental Setup in Full}
\label{app:setup}

\paragraph{Models and scope}
GPT-2 small \citep{radford2019gpt2} (124M, $d{=}768$, 12 blocks) and Mistral-7B-v0.1 \citep{jiang2023mistral} ($56\times$ larger, 32 layers, $d{=}4096$), which additionally uses grouped-query attention, rotary embeddings and a sliding-window mask. In both, \emph{only} the QKV projection is quantized, so absolute perplexities are not comparable to end-to-end quantized systems.

\paragraph{Runs}
Three runs are reported, all using $\texttt{group\_size}{=}128$, \texttt{act\_order}, $\lambda_{\text{KL}}{=}0.1$ unless ablated, and 10 Hutchinson probes on 2 calibration batches per layer. \textbf{Run A} calibrates and evaluates GPT-2 on WikiText-2 and sweeps criterion $\times$ 7 budgets $\times$ \{GPTQ, ${+}$FT\}. \textbf{Run B} calibrates GPT-2 on C4 and evaluates on \emph{both} corpora, sweeping 5 modes $\times$ 5 budgets under all four metrics. \textbf{Run C} sweeps the same 5 modes $\times$ 5 budgets on Mistral-7B in all four WikiText-2/C4 calibrate$\to$evaluate pairings, plus a sensitivity-source $\times$ allocator cross. Mistral-7B (14.5\,GB in fp16) does not fit on a free-tier T4 alongside the float teacher copy the fine-tuning stage requires, so Run C uses a \emph{layer-streaming} re-implementation: shards are fetched tensor-by-tensor onto the GPU and each decoder layer is materialized on the \texttt{meta} device and discarded after use, at the cost of two passes --- score, then quantize-and-evaluate --- per configuration. This is why Run C is a re-implementation rather than a re-run.

\paragraph{Evaluation corpora}
Runs A and B score GPT-2 on the full WikiText-2 \citep{merity2017pointer} test split; Run B additionally scores the \emph{same} weights in-domain on a disjoint C4 \citep{raffel2020c4} validation sample, which is what lets calibration effects be separated from evaluation effects in Sections~\ref{sec:ft} and \ref{sec:cross}. Since C4 has no canonical evaluation subset, those absolute perplexities are specific to this draw; and Run C uses 32 windows ($\approx$33k tokens) per configuration, so its perplexities are subset figures --- exact within a pairing, not comparable to published Mistral-7B numbers --- with $\varepsilon^W,\varepsilon^A$ on the streaming pipeline's raw loss scale.

\section{The Sensitivity Criterion}
\label{sec:sens}

\begin{figure*}[!t]
\centering
\includegraphics[width=0.64\textwidth]{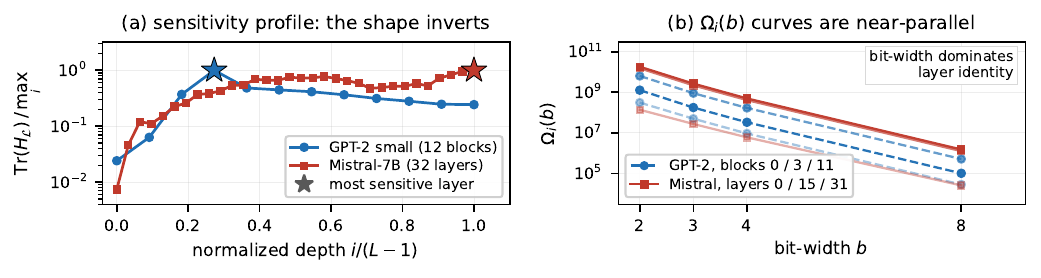}
\caption{Sensitivity criterion across both models. (a) Hessian trace per layer, normalized by each model's maximum, against normalized depth; stars mark the most sensitive layer. The profile is strongly non-uniform in both models ($41\times$ spread on GPT-2, $135\times$ on Mistral) but its shape \emph{inverts} with architecture. (b) The sensitivity tables $\Omega_i(b)$ of \eqref{eq:omega} are near-parallel in both models --- a shared, steep sub-4-bit cliff --- so bit-width dominates layer identity by orders of magnitude.}
\label{fig:sens}
\end{figure*}

\paragraph{Sensitivity is strongly non-uniform, but its depth profile is architecture-specific}
All traces are positive, as \eqref{eq:gn} requires, and both models span one to two orders of magnitude across layers ($41\times$ on GPT-2, $135\times$ on Mistral), so the flat profile a uniform allocation implicitly assumes is clearly wrong --- the QKV analogue of the head-level heterogeneity reported by \citet{michel2019heads}. The \emph{shape}, however, does not transfer (Fig.~\ref{fig:sens}(a)): GPT-2 peaks at block 3 of 12 and decays, whereas Mistral rises with depth and peaks at its final layers. This falsifies C2 as a general rule, since the $(L-i)$ weighting of \eqref{eq:omega} protects \emph{early} layers --- which rescues GPT-2 at 3 bits (Section~\ref{sec:allocres}) but at Mistral scale would suppress exactly the layers the criterion ranks highest.

\paragraph{Bit-width dominates layer identity}
The per-layer $\Omega_i(b)$ curves are near-parallel in both models (Fig.~\ref{fig:sens}(b)): dropping from 4 to 3 bits multiplies $\Omega$ by roughly $5\times$ and from 3 to 2 by a further $5$--$7\times$, almost independently of which layer is scored. This shared cliff is a property of the construction, not of one model, and it means a ratio-based allocator has little to differentiate layers on until the budget forces one below 4 bits --- where the choice then matters a great deal.

\paragraph{The ranking is a property of the model, not of the calibration text}
Traces computed under WikiText-2 (Run A) and C4 (Run B) calibration correlate at $r = 0.9995$ and induce the same ordering up to one adjacent pair differing by under $1\%$ --- not obvious, since both the Hessian and the perturbation term of \eqref{eq:omega} are activation-dependent.

\paragraph{The score matters; the solver does not}
A better sensitivity estimate pays off only where allocation is genuinely hard, and the choice of solver hardly matters at all. Crossing the two sensitivity sources with the two allocators on Mistral (Wiki$\to$Wiki, joint FT throughout), second-order Hutchinson scoring beats the first-order Fisher proxy at the aggressive 3-bit budget ($5.268$ vs.\ $5.302$ perplexity, $0.6090$ vs.\ $0.6071$ accuracy), whereas greedy and MCKP/Pareto given identical scores are indistinguishable there, and all four combinations lie within $0.007$ perplexity at 6 bits. This mirrors the criterion-versus-oracle result of Section~\ref{sec:allocres}: the double-backward cost is worth paying, the Pareto pre-filter is not.

\section{Bit Allocation}
\label{sec:allocres}

\begin{figure*}[!t]
\centering
\includegraphics[width=0.80\textwidth]{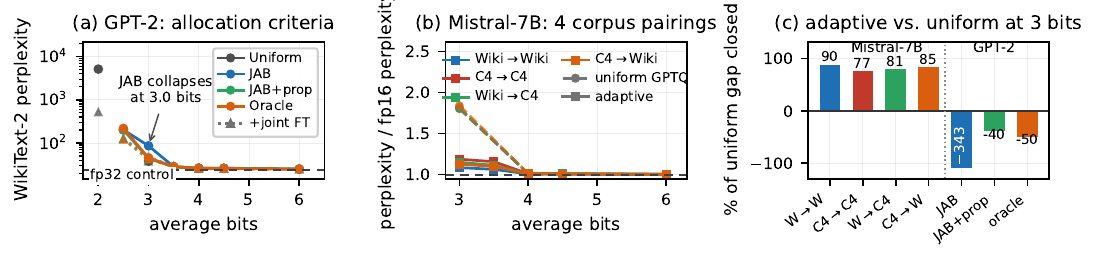}
\caption{Bit allocation on the QKV projection. (a) GPT-2: perplexity vs.\ average bits for four
criteria, GPTQ-only (solid) and ${+}$joint fine-tuning (dotted). (b) Mistral-7B: perplexity
normalized by each pairing's fp16 baseline, uniform GPTQ vs.\ Hutchinson-greedy adaptive.
(c) Share of the uniform-GPTQ gap recovered by adaptive allocation at 3 bits: Mistral recovers
$77$--$90\%$ in every pairing, while on GPT-2 the same criterion \emph{loses} $343\%$ (bar clipped),
reduced to $-40\%$ by the depth re-weighting C2 and $-50\%$ by the oracle.}
\label{fig:alloc}
\end{figure*}

\begin{table}[!t]
\centering
\caption{GPT-2 Run A: full WikiText-2 test perplexity across criteria, budgets, and modes. fp32 control \textbf{24.357}. ``--'' marks combinations that do not exist. Best per row in \textbf{bold}.}
\label{tab:sweep}
\scriptsize
\setlength{\tabcolsep}{2.4pt}
\begin{tabular}{c cc cc cc cc}
\toprule
& \multicolumn{2}{c}{Uniform} & \multicolumn{2}{c}{JAB (C1)} & \multicolumn{2}{c}{JAB${+}$prop (C2)} & \multicolumn{2}{c}{Oracle (C3)} \\
\cmidrule(lr){2-3}\cmidrule(lr){4-5}\cmidrule(lr){6-7}\cmidrule(lr){8-9}
bits & GPTQ & ${+}$FT & GPTQ & ${+}$FT & GPTQ & ${+}$FT & GPTQ & ${+}$FT \\
\midrule
2.0 & 5137.6 & \textbf{516.8} & -- & -- & -- & -- & -- & -- \\
2.5 & -- & -- & 201.7 & 126.5 & 201.7 & 126.5 & 218.0 & \textbf{120.1} \\
3.0 & 38.35 & \textbf{37.81} & 86.36 & 48.31 & 43.93 & 41.99 & 45.36 & 42.45 \\
3.5 & -- & -- & \textbf{28.32} & 28.38$^{\dagger}$ & \textbf{28.32} & 28.38$^{\dagger}$ & 28.61 & 29.09 \\
4.0 & 26.23 & \textbf{25.85} & 26.23 & \textbf{25.85} & 26.23 & \textbf{25.85} & 26.23 & \textbf{25.85} \\
4.5 & -- & -- & 26.20 & \textbf{25.62} & 26.20 & \textbf{25.62} & 25.87 & 25.68 \\
6.0 & -- & -- & 25.05 & \textbf{24.79} & 25.05 & \textbf{24.79} & 25.11 & 24.80 \\
\bottomrule
\multicolumn{9}{l}{\footnotesize $^{\dagger}$ the only budget at which fine-tuning hurts; see Section~\ref{sec:ft}.}
\end{tabular}
\end{table}

\paragraph{Mixed precision helps above 4 bits, ties at 4, and can be catastrophic below}
On GPT-2 (Table~\ref{tab:sweep}, Fig.~\ref{fig:alloc}(a)) adaptive allocation improves on uniform 4-bit by $-1.18$ perplexity at 6 average bits by upgrading high-trace blocks to 8 bits. At exactly 4.0 bits it returns an all-4-bit assignment and reproduces the uniform baseline exactly --- correct, since a budget the uniform solution already saturates leaves nothing to trade. At 3.0 bits it instead reaches $86.36$ against uniform 3-bit's $38.35$: the allocator is destroying the model.

\paragraph{That failure is a provable consequence of the additive objective}
Suppose the 3.0-bit assignment contained no block below 3 bits. Then $\sum_i b_i \ge 36$, and the budget caps $\sum_i b_i \le 36$, so equality would force $b_i=3$ everywhere --- the uniform assignment and its 38.345. It does not, so the allocation must contain at least one 2-bit block, which Run B confirms by inspection (blocks 0 and 1). One such block suffices: uniform 2-bit GPTQ scores 5137.6, over $200\times$ the fp32 control. Minimizing a \emph{sum} of $\Omega_i(b_i)$, the allocator cannot see that perplexity is convex and effectively unbounded in the low-bit direction, so the downgrade looks ratio-favourable in \eqref{eq:omega} and is not. Consistent with this, the depth re-weighting C2 --- which costs nothing and leaves every other budget's assignment unchanged --- halves the damage ($43.93$ vs.\ $86.36$) simply by making the sacrifice of an early block look expensive.

\paragraph{Mixed precision earns its keep at budgets uniform cannot express}
Its real value is at fractional budgets, where no uniform alternative exists: at 2.5 and 3.5 average bits it beats a geometric interpolation between the neighbouring uniform solutions by $2.2\times$ and $10.7\%$.

\paragraph{A cheap proxy is competitive with a 73-model-load oracle}
The oracle C3 costs $\approx73\times$ the model instantiations of C1 and buys little: across the four budgets $\geq3.5$ bits the mean absolute difference is $0.17$ perplexity, with the cheap criterion ahead in half the cases. C3 helps decisively only at 3.0 bits, where it fixes the same 2-bit pathology the free depth re-weighting largely fixes --- and neither fixes it as well as declining to use mixed precision there. Above 3.5 bits the allocation is essentially solved, and the residual gap to fp32 lies in the quantizer.

\paragraph{At 7B scale allocation carries the result, and the 3-bit collapse does not reproduce}
Adaptive allocation dominates uniform GPTQ at 3 bits in all four corpus pairings (Table~\ref{tab:mistral}), recovering $77$--$90\%$ of the uniform-versus-fp16 perplexity gap (Fig.~\ref{fig:alloc}(c)) and $2$--$3$ accuracy points; by 4 bits every method is within $1$--$2\%$ of fp16 and by 6 bits within $0.5\%$, the same diminishing-returns pattern as GPT-2. What does \emph{not} appear is a 3-bit collapse, even though the budget arithmetic applies equally: Mistral's allocator must also place some layer below 3 bits. The per-layer assignment is unavailable, so we report this as an open discrepancy. Two untested explanations are plausible: with 32 blocks rather than 12, one low-bit layer is a much smaller share of the budget; and GQA shrinks the K/V projections, changing both $c(b)$ and the curvature landscape.

\begin{table*}[!t]
\centering
\caption{Mistral-7B, all four calibrate$\to$evaluate corpus pairings. Upper block: perplexity (lower better). Lower block: next-token top-1 accuracy (higher better). fp16 baselines: PPL 4.818 (WikiText-2 eval) / 7.492 (C4 eval); Acc 0.6257 / 0.5500. ``--'' marks a flat-bit mode at a fractional budget, left undefined rather than silently rounded. Best per column in \textbf{bold}; ties bolded.}
\label{tab:mistral}
\scriptsize
\setlength{\tabcolsep}{2.6pt}
\resizebox{0.88\textwidth}{!}{%
\begin{tabular}{l ccccc ccccc ccccc ccccc}
\toprule
& \multicolumn{5}{c}{Wiki$\to$Wiki} & \multicolumn{5}{c}{C4$\to$C4} & \multicolumn{5}{c}{Wiki$\to$C4} & \multicolumn{5}{c}{C4$\to$Wiki} \\
\cmidrule(lr){2-6}\cmidrule(lr){7-11}\cmidrule(lr){12-16}\cmidrule(lr){17-21}
Method & 3b & 3.5b & 4b & 4.5b & 6b & 3b & 3.5b & 4b & 4.5b & 6b & 3b & 3.5b & 4b & 4.5b & 6b & 3b & 3.5b & 4b & 4.5b & 6b \\
\midrule
\multicolumn{21}{l}{\textit{Perplexity}} \\
Uniform (GPTQ)                  & 8.738 & -- & 4.855 & -- & \textbf{4.822} & 13.506 & -- & \textbf{7.571} & -- & \textbf{7.495} & 13.575 & -- & \textbf{7.580} & -- & \textbf{7.496} & 8.875 & -- & 4.870 & -- & \textbf{4.820} \\
Adaptive (Hutch., greedy)       & \textbf{5.228} & 5.129 & 4.855 & 4.850 & 4.834 & 8.898 & 8.670 & \textbf{7.571} & \textbf{7.556} & 7.518 & 8.622 & 8.340 & \textbf{7.580} & 7.577 & 7.530 & 5.428 & 5.302 & 4.870 & 4.861 & 4.839 \\
Joint (uniform bits)            & 8.833 & -- & 4.875 & -- & \textbf{4.822} & 13.615 & -- & 7.617 & -- & 7.520 & 13.658 & -- & 7.624 & -- & \textbf{7.496} & 8.855 & -- & 4.881 & -- & 4.821 \\
JAB${+}$Joint (Hutch., greedy)  & 5.268 & 5.134 & 4.875 & 4.867 & 4.845 & \textbf{8.732} & \textbf{8.507} & 7.617 & 7.595 & 7.553 & 8.681 & 8.373 & 7.624 & 7.608 & 7.561 & \textbf{5.411} & \textbf{5.268} & 4.881 & 4.873 & 4.843 \\
JAB${+}$Joint, no KL            & 5.231 & \textbf{5.121} & \textbf{4.854} & \textbf{4.849} & 4.835 & 8.805 & 8.563 & 7.575 & 7.559 & 7.539 & \textbf{8.582} & \textbf{8.337} & 7.583 & \textbf{7.575} & 7.531 & 5.477 & 5.293 & \textbf{4.866} & \textbf{4.858} & 4.838 \\
\midrule
\multicolumn{21}{l}{\textit{Accuracy}} \\
Uniform (GPTQ)                  & 0.5786 & -- & 0.6236 & -- & \textbf{0.6252} & 0.5094 & -- & \textbf{0.5478} & -- & \textbf{0.5504} & 0.5086 & -- & 0.5486 & -- & 0.5498 & 0.5752 & -- & 0.6230 & -- & \textbf{0.6256} \\
Adaptive (Hutch., greedy)       & 0.6093 & \textbf{0.6146} & 0.6236 & 0.6248 & 0.6249 & 0.5229 & 0.5254 & \textbf{0.5478} & \textbf{0.5485} & 0.5498 & 0.5269 & \textbf{0.5309} & 0.5486 & 0.5483 & \textbf{0.5499} & 0.6034 & 0.6084 & 0.6230 & 0.6235 & 0.6253 \\
Joint (uniform bits)            & 0.5762 & -- & 0.6230 & -- & \textbf{0.6252} & 0.5066 & -- & 0.5470 & -- & 0.5494 & 0.5067 & -- & 0.5475 & -- & 0.5498 & 0.5731 & -- & 0.6232 & -- & 0.6254 \\
JAB${+}$Joint (Hutch., greedy)  & 0.6090 & 0.6128 & 0.6230 & 0.6233 & 0.6238 & \textbf{0.5245} & \textbf{0.5286} & 0.5470 & 0.5484 & 0.5493 & 0.5269 & 0.5298 & 0.5475 & 0.5480 & 0.5489 & \textbf{0.6058} & \textbf{0.6100} & 0.6232 & \textbf{0.6239} & 0.6247 \\
JAB${+}$Joint, no KL            & \textbf{0.6112} & 0.6145 & \textbf{0.6240} & \textbf{0.6249} & 0.6245 & \textbf{0.5245} & \textbf{0.5286} & 0.5473 & 0.5483 & 0.5493 & \textbf{0.5272} & 0.5307 & \textbf{0.5488} & \textbf{0.5489} & \textbf{0.5499} & 0.6033 & 0.6079 & \textbf{0.6235} & 0.6226 & 0.6246 \\
\bottomrule
\end{tabular}%
}
\end{table*}

\section{Quantization Method: GPTQ vs.\ Joint Fine-Tuning}
\label{sec:ft}

\begin{figure*}[!t]
\centering
\includegraphics[width=0.83\textwidth]{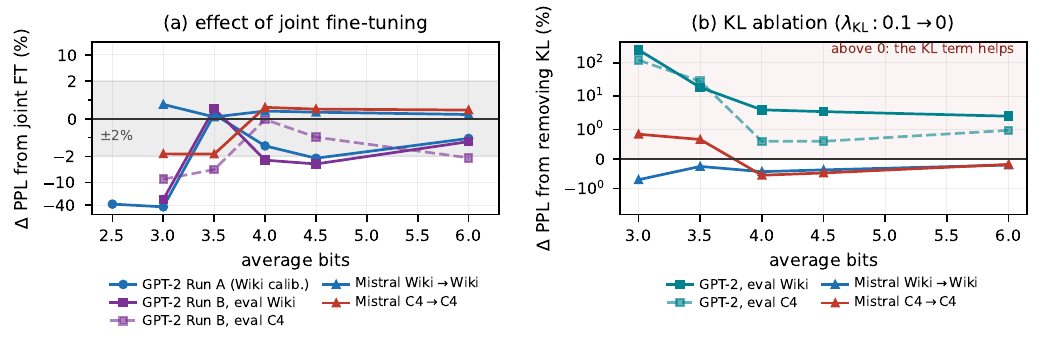}
\caption{Joint fine-tuning across models, corpora and budgets, on symmetric-log axes so that both the low-bit and near-lossless regimes are legible. (a) Relative perplexity change from adding the joint STE stage to the same allocation; negative is better, shaded band $\pm2\%$. Gains are large exactly where quantization damage is large and shrink to a few percent in the usable regime, where the sign becomes model- and corpus-dependent. (b) The KL ablation, $\lambda_{\text{KL}}: 0.1\to0$: on GPT-2 removing the term is harmful at every budget on both corpora, catastrophically so below 4 bits; on Mistral, where the stage moves far less, the effect is within $\pm1\%$ and changes sign.}
\label{fig:ft}
\end{figure*}

\paragraph{Fine-tuning helps most where quantization damage is largest}
On GPT-2 the joint stage improves 18 of the 21 configurations of Table~\ref{tab:sweep}. Gains scale with the damage --- $-45\%$ to $-90\%$ below 3 bits --- and settle at $1$--$2\%$ in the usable regime, bringing the best arm (JAB at 6 average bits) to $24.786$ against an fp32 control of $24.357$. A flip rate of $0.12$--$0.18$ throughout confirms this reflects the deployed weights, not latent shadow weights.

\paragraph{Its one failure is a proxy--objective gap the commit gate cannot see}
All three exceptions sit at 3.5 average bits, where the gate of Section~\ref{sec:ste}(4) certifies an improvement on the held-out attention loss while perplexity worsens slightly. There the allocation holds only 3- and 4-bit blocks, so every block carries substantial error and none is nearly exact; the optimizer has room for large local improvements, some of which trade against the global objective.

\paragraph{Fine-tuning's instabilities are corpus effects, and they separate cleanly}
Scoring the identical Run B weights on both corpora isolates cause from symptom. The 3.5-bit anomaly is an \emph{evaluation}-corpus artifact: in-domain on C4 the sign reverses ($-4.5\%$ instead of $+0.5\%$) although nothing about the models changed, so the agreement between Runs A and B here is better read as two measurements sharing an evaluation corpus than as independent confirmation --- a caution for any single-corpus ablation. The 3-bit sign flip against Run A is instead a \emph{calibration}-corpus effect: moving to the matched corpus makes it worse ($+60.8\%$ versus $+16.1\%$), so cross-corpus evaluation cannot be the cause. Both are confined to fractional and aggressive budgets.

\paragraph{At 7B scale, refinement is second-order to allocation}
Fig.~\ref{fig:ft}(a) places every fine-tuning comparison on one axis. On Mistral the correction is marginally negative on WikiText-2 and marginally positive on C4; the magnitudes are what matter, since at 3 bits allocation is worth $3$--$5$ perplexity points and the fine-tuning at most $0.17$ either way. From an independent model and codebase this reinforces the central claim --- \emph{how bits are spent matters more than how the resulting weights are locally refined} --- subject to two caveats: the LSQ and grid-relative refinements were not re-verified at this scale, and the streaming build's smaller step budget makes its fine-tuning stage less comparable to GPT-2's than its allocator is.

\paragraph{The KL term is what makes fine-tuning work}
Holding bit-widths fixed on GPT-2, removing the attention-map term recovers only $-0.052$ of the $-0.444$ total gain, so the KL term supplies $88\%$ of it --- consistent with the reduction imbalance of Section~\ref{sec:loss} making the objective effectively attention-map distillation. Fig.~\ref{fig:ft}(b) sweeps the ablation: dropping the term is worse than keeping it at every budget on \emph{both} evaluation corpora, catastrophically so below 4 bits. The mechanism is visible in the metrics rather than assumed. MSE-only fine-tuning \emph{does} reduce attention-output error relative to no fine-tuning ($\varepsilon^{A}: 0.238 \to 0.188$ at 4 bits) --- it succeeds at the objective it was given --- yet perplexity still worsens. Matching the attention \emph{output} is therefore not sufficient: an unconstrained MSE descent can reach a lower output error through a distribution the downstream stack was never trained against, and the KL term forbids that.

\paragraph{Flip rate measures movement, not benefit}
At a fixed assignment the MSE-only run moves \emph{more} grid positions than the KL run ($0.177$ vs.\ $0.136$) while ending $0.392$ perplexity worse, so activity and benefit must be logged separately. Benefit is monotone in fine-tuning depth --- restricting the stage to the first four blocks lands between GPTQ-only and full-depth --- which refutes a compounding-drift hypothesis: had each block accumulated error against a stale target, a prefix would have beaten full depth.

\section{Where the Error Lives, and Whether Accuracy Agrees}
\label{sec:errspace}

\begin{table*}[!t]
\centering
\caption{GPT-2 Run B (C4 calibration): five pipeline modes, five budgets, four metrics, evaluated on both corpora. $\varepsilon^{W}$ is a property of the quantized weights alone and is therefore shared; the amplification factor $\rho=\overline{\varepsilon^{A}}/\overline{\varepsilon^{W}}$ is plotted in Fig.~\ref{fig:err}(b). Controls: WikiText-2 24.357 / 0.4148; C4 30.778 / 0.3786. ``--'' marks a flat-bit mode at a fractional budget.}
\label{tab:runb}
\tiny
\setlength{\tabcolsep}{4pt}
\begin{tabular}{l c ccc ccc c}
\toprule
& & \multicolumn{3}{c}{Evaluated on WikiText-2 (cross-corpus)} & \multicolumn{3}{c}{Evaluated on C4 (in-domain)} & \\
\cmidrule(lr){3-5}\cmidrule(lr){6-8}
Mode & bits & PPL & Acc & $\varepsilon^{A}$ & PPL & Acc & $\varepsilon^{A}$ & $\varepsilon^{W}$ \\
\midrule
uniform & 3   & 39.338 & 0.3612 & 0.4711 & 54.862 & 0.3339 & 0.4621 & 0.3753 \\
uniform & 4   & 26.536 & 0.4038 & 0.2382 & 33.638 & 0.3692 & 0.2370 & 0.1739 \\
uniform & 6   & 24.488 & 0.4140 & 0.0695 & 30.955 & 0.3774 & 0.0702 & 0.0409 \\
\midrule
jab & 3   & 80.140 & 0.2924 & 0.4463 & 99.206 & 0.2751 & 0.4287 & 0.4321 \\
jab & 3.5 & 29.425 & 0.3949 & 0.3579 & 36.958 & 0.3608 & 0.3504 & 0.2787 \\
jab & 4   & 26.536 & 0.4038 & 0.2382 & 33.631 & 0.3692 & 0.2370 & 0.1739 \\
jab & 4.5 & 26.315 & 0.4055 & 0.2206 & 33.348 & 0.3703 & 0.2193 & 0.1591 \\
jab & 6   & 25.042 & 0.4107 & 0.1178 & 31.528 & 0.3751 & 0.1145 & 0.0908 \\
\midrule
joint & 3   & 45.653 & 0.3456 & 0.3208 & \textbf{88.226} & \textbf{0.3182} & 0.2912 & 0.4595 \\
joint & 4   & 25.867 & 0.4078 & 0.1602 & \textbf{33.347} & \textbf{0.3705} & 0.1472 & 0.2075 \\
joint & 6   & \textbf{24.465} & \textbf{0.4143} & 0.0440 & 30.953 & 0.3773 & 0.0414 & 0.0492 \\
\midrule
adaptive\_joint & 3   & 57.753 & 0.3188 & 0.3135 & 91.149 & 0.2914 & 0.2801 & 0.5192 \\
adaptive\_joint & 3.5 & 29.585 & 0.3922 & 0.2475 & \textbf{35.301} & 0.3598 & 0.2246 & 0.3313 \\
adaptive\_joint & 4   & \textbf{25.867} & \textbf{0.4078} & 0.1602 & 33.619 & \textbf{0.3705} & 0.1472 & 0.2075 \\
adaptive\_joint & 4.5 & \textbf{25.478} & \textbf{0.4105} & 0.1474 & \textbf{33.022} & \textbf{0.3717} & 0.1348 & 0.1899 \\
adaptive\_joint & 6   & 24.738 & 0.4138 & 0.0819 & \textbf{30.838} & 0.3757 & 0.0748 & 0.1064 \\
\midrule
kl0 ($\lambda_{\text{KL}}{=}0$) & 3   & 199.247 & 0.2052 & 0.4461 & 200.900 & 0.1970 & 0.4157 & 0.5554 \\
kl0 & 3.5 & 34.980 & 0.3703 & 0.3062 & 45.137 & 0.3417 & 0.2819 & 0.3599 \\
kl0 & 4   & 26.865 & 0.4017 & 0.1883 & 33.819 & 0.3690 & 0.1712 & 0.2167 \\
kl0 & 4.5 & 26.348 & 0.4050 & 0.1749 & 33.219 & 0.3707 & 0.1593 & 0.1978 \\
kl0 & 6   & 25.353 & 0.4099 & 0.0996 & 31.136 & 0.3750 & 0.0908 & 0.1133 \\
\bottomrule
\end{tabular}
\end{table*}

\begin{figure*}[!t]
\centering
\includegraphics[width=0.91\textwidth]{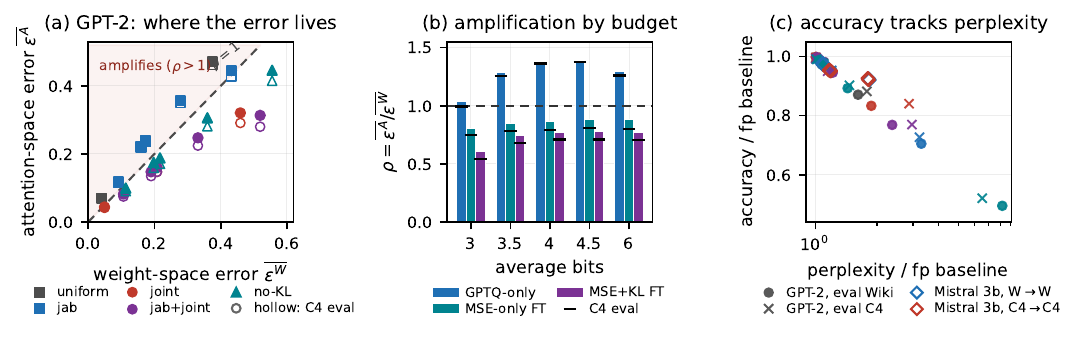}
\caption{(a) Attention-space against weight-space error, one point per GPT-2 Run B arm; filled markers WikiText-2 eval, hollow markers C4 eval. Every GPTQ-only arm sits in the shaded amplifying region above the $\rho=1$ diagonal, every fine-tuned arm below it. (b) The same as the amplification factor $\rho$ by budget: the ordering GPTQ-only $>$ MSE-only $>$ MSE${+}$KL holds at every budget on both corpora without a crossing. (c) Accuracy against perplexity, each normalized by its own full-precision baseline, pooling GPT-2 Run B with the Mistral-7B 3-bit arms.}
\label{fig:err}
\end{figure*}

\paragraph{Fine-tuning trades weight fidelity for attention fidelity}
This measurement most directly tests the report's premise, and the premise survives. On the flat 4-bit pair, joint fine-tuning moves the weights $19\%$ \emph{further} from float ($\varepsilon^{W}: 0.174 \to 0.208$) while moving the attention output $33\%$ \emph{closer} ($\varepsilon^{A}: 0.238 \to 0.160$), and both perplexity and accuracy improve; the same trade appears at 3 and 6 bits. A solution measurably worse by the objective GPTQ minimizes is therefore better on the metric that matters and better on the task, so the weight-space objective is not the binding constraint.

\paragraph{The amplification factor separates the two regimes without exception}
Causal softmax attention amplifies the relative weight perturbation it is handed, and fine-tuning inverts that: every GPTQ-only arm has $\rho>1$ and every fine-tuned arm $\rho<1$, with the ordering $\rho_{\text{GPTQ}}>\rho_{\text{MSE-only}}>\rho_{\text{MSE+KL}}$ holding at every budget on both corpora without a crossing (Fig.~\ref{fig:err}(a,b)). The single departure --- \texttt{jab} at 3 bits, where $\rho$ moves from $1.03$ to $0.99$ between corpora --- is smaller than single-draw Hutchinson estimates at one seed can resolve.

\paragraph{On Mistral the same dissociation is produced by the allocator instead}
At 3 bits, adaptive allocation attains a lower attention-reconstruction error than uniform on both calibration corpora ($1.13$ vs.\ $1.53\times10^{-3}$ under WikiText-2 calibration, $1.32$ vs.\ $1.65\times10^{-3}$ under C4) while incurring a \emph{higher} weight error ($0.467$ vs.\ $0.433$ and $0.474$ vs.\ $0.439$) --- exactly what spending bits on high-curvature layers implies --- and again it is that trade, not weight fidelity, that tracks perplexity and accuracy. The same signature arising through two mechanisms in two independent codebases is stronger evidence than either alone. (Both errors are on the streaming pipeline's raw scale.)

\paragraph{Accuracy confirms the perplexity ordering}
Next-token accuracy is almost perfectly rank-inverse to perplexity across all 21 GPT-2 arms (Spearman $\rho_s = -0.996$ on WikiText-2, $-0.990$ on C4), and Fig.~\ref{fig:err}(c) shows the relation holding across models, corpora and modes once each configuration is normalized by its own full-precision baseline --- so the differences above are not artifacts of the likelihood tail. Only two arms reorder by more than one rank, both at budgets already flagged as unstable.

\section{Cross-Corpus Robustness}
\label{sec:cross}

\paragraph{Quantization cost is a property of calibration and budget, not of the scoring corpus}
Uniform 4-bit GPTQ costs $+8.9\%$ perplexity against the WikiText-2 control and $+9.3\%$ against the C4 control despite baselines differing by 6.4 absolute points.

\paragraph{Allocation transfers across corpus mismatch; refinement does not}
On Mistral, adaptive allocation recovers $81.4\%$ and $85.0\%$ of the uniform gap under corpus mismatch, against $89.5\%$ and $76.6\%$ matched --- so the mismatch penalty is smaller than the spread \emph{between} the two matched pairings (Fig.~\ref{fig:alloc}(c)). The sensitivity ranking is therefore a property of the model's architecture and weight structure rather than an artifact overfit to the calibration text: the functional counterpart of the $r=0.9995$ trace correlation measured directly on GPT-2. Fine-tuning behaves oppositely --- both of its instabilities are corpus-induced (Section~\ref{sec:ft}) --- while no allocation result changes sign under either intervention.

\section{Beyond QKV: the MLP Sublayer}
\label{sec:mlp}

The pipeline of Section~\ref{sec:method} is not specific to attention: $\Ljab$ is a block-level reconstruction loss, and nothing in the Hutchinson estimator, the $\Omega_i(b)$ table or the allocator refers to a softmax. We therefore ported the whole system to the MLP sublayer, quantizing $\texttt{c\_fc}$ and $\texttt{mlp.c\_proj}$ and leaving all attention weights in fp32. Two changes are structural. First the MLP is \emph{two sequential} matrices rather than one fused one, so each block contributes two allocation units and two Hessians, of dimension $d$ and $4d$; the second is collected only after the first is quantized, preserving within-block the sequential property Section~\ref{sec:gptq} obtains across blocks. Second, and decisively, \emph{the attention map has no MLP analogue}: $\mathcal{L}_{\text{KL}}$ cannot be formed, so $\lambda_{\text{KL}}=0$ by construction and the objective reduces to output MSE alone. The KL ablation of Section~\ref{sec:ft} is correspondingly vacuous here, and we verified it reproduces the default arm bit-for-bit ($\Delta=0.0\mathrm{e}{+}00$).

The scope is worth stating, because it inverts. The two MLP matrices hold $8d^2$ parameters per block against QKV's $3d^2$: $56.6$M versus $21.2$M on GPT-2, or $66.6\%$ of non-embedding weights against $25.0\%$. Every result below therefore covers $2.67\times$ more of the model than Table~\ref{tab:sweep}, and no MLP budget is reachable by any QKV-only scheme --- even 0-bit QKV leaves a larger model than 4-bit MLP. Compression at 4 bits is $1.66\times$ end-to-end, the residue being the $39.4$M embedding parameters ($32\%$ of GPT-2 small) that no arm in this report touches.

\begin{table}[!t]
\centering
\caption{GPT-2 MLP: full WikiText-2 test perplexity, same pipeline, criteria and budgets as Table~\ref{tab:sweep}, with $\texttt{c\_fc}$ and $\texttt{mlp.c\_proj}$ quantized and attention left in fp32. fp32 control \textbf{24.357}; budget is average bits per matrix over 24 units. ``--'' marks combinations that do not exist. Best per row in \textbf{bold}. Contrast the Uniform column with Table~\ref{tab:sweep}: here the criterion \emph{loses} to it at both comparable budgets.}
\label{tab:mlp}
\scriptsize
\setlength{\tabcolsep}{2.4pt}
\begin{tabular}{c cc cc cc cc}
\toprule
& \multicolumn{2}{c}{Uniform} & \multicolumn{2}{c}{JAB (C1)} & \multicolumn{2}{c}{JAB${+}$prop (C2)} & \multicolumn{2}{c}{Oracle (C3)} \\
\cmidrule(lr){2-3}\cmidrule(lr){4-5}\cmidrule(lr){6-7}\cmidrule(lr){8-9}
bits & GPTQ & ${+}$FT & GPTQ & ${+}$FT & GPTQ & ${+}$FT & GPTQ & ${+}$FT \\
\midrule
2.0 & \textbf{3698.3} & 8462.0$^{\ddagger}$ & -- & -- & -- & -- & -- & -- \\
2.5 & -- & -- & 2458.3 & 1439.6 & 395.7 & 248.2 & 351.3 & \textbf{275.1} \\
3.0 & 53.71 & 46.45 & 120.70 & 94.49 & 53.71 & 46.45 & 46.15 & \textbf{39.13} \\
3.5 & -- & -- & 40.12 & 37.55 & 29.90 & 28.80 & 28.54 & \textbf{27.99} \\
4.0 & 26.23 & \textbf{25.96} & 27.46 & 26.64 & 26.23 & \textbf{25.96} & 26.23 & \textbf{25.96} \\
4.5 & -- & -- & 26.14 & 25.89 & 26.11 & 25.76 & 25.60 & \textbf{25.43} \\
6.0 & -- & -- & 25.68 & 25.20 & 24.92 & 24.98$^{\ddagger}$ & 24.76 & \textbf{24.68} \\
\bottomrule
\multicolumn{9}{l}{\footnotesize $^{\ddagger}$ the two budgets at which fine-tuning hurts.}
\end{tabular}
\end{table}

\paragraph{The criterion loses to uniform, and the oracle does not}
Two budgets admit a uniform baseline, and C1 is beaten at both: $120.70$ against $53.71$ at 3 bits ($2.25\times$ worse) and $27.46$ against $26.23$ at 4 bits. This is not the 3.0-bit pathology of Section~\ref{sec:alloc}, whose cause was a 2-bit block entering the allocation; here the 4-bit assignment contains nothing below 3 bits and still loses. Adaptive allocation as implemented is simply harmful on this sublayer. That the oracle C3 \emph{beats} uniform at 3 bits ($46.15$ vs.\ $53.71$, a $14\%$ improvement) establishes that exploitable structure exists and that the Hutchinson criterion is failing to find it --- a cleaner separation of criterion from ceiling than the QKV sweep affords, where C1 and C3 agree to $0.17$ perplexity above 3.5 bits.

\paragraph{The depth profile inverts again, and is far flatter}
Fig.~\ref{fig:sens}(a) reported a QKV trace peaking at block 3 of 12 and decaying, with a $41\times$ spread. The MLP trace on the \emph{same model} rises monotonically from block 5 to the last layer and spans only $4.5\times$ ($365\to1543$). ``Architecture-specific'', the reading offered in Section~\ref{sec:sens}, is therefore too coarse: the profile is \emph{sublayer}-specific, and inverts between two sublayers of one block. This weakens the $(L-i)$ re-weighting of \eqref{eq:omega} further than Fig.~\ref{fig:sens}(a) already does --- C2's advantage over C1 here (e.g.\ $29.90$ vs.\ $40.12$ at 3.5 bits) comes from opposing a depth gradient it was designed to create.

\paragraph{The trace cannot separate the two matrices, by construction}
A single Hutchinson trace is taken over the concatenated $[W_{\text{fc}} \mid W_{\text{proj}}]$ vector, so $\mathrm{Tr}(H)$ is identical for the two units of a block and all discrimination between them falls to the perturbation factor of \eqref{eq:omega}. This is the exact MLP analogue of Section~\ref{sec:joint}'s observation that the trace sees only diagonal blocks: joint and separate coincide for scoring. Here that identity is not benign, because --- unlike $Q$, $K$ and $V$ --- the two MLP matrices have different input dimensions, different input distributions, and, as the oracle shows, genuinely different sensitivities.

\begin{table}[t]
\centering
\caption{GPT-2 MLP, bit assignments at the 3.5-bit budget by unit and depth. The oracle's rule is
constant in depth and splits by matrix role; the Hutchinson criterion follows a depth gradient
instead and pays $11.6$ perplexity for it.}
\label{tab:mlpalloc}
\small
\setlength{\tabcolsep}{4pt}
\begin{tabular}{@{}l l l c@{}}
\toprule
Criterion & Unit & Assignment (blocks 0--11) & PPL \\
\midrule
Oracle (C3) & \texttt{c\_fc}   & \texttt{4 4 4 4 4 4 4 4 4 4 4 4} & \multirow{2}{*}{\textbf{28.54}} \\
            & \texttt{c\_proj} & \texttt{3 3 3 3 3 3 3 3 3 3 3 3} & \\
\addlinespace
JAB (C1)    & \texttt{c\_fc}   & \texttt{4 3 3 3 3 3 4 4 4 4 4 4} & \multirow{2}{*}{40.12} \\
            & \texttt{c\_proj} & \texttt{3 3 3 3 3 3 3 4 4 4 4 4} & \\
\bottomrule
\end{tabular}
\end{table}

\paragraph{The oracle finds a role rule, not a depth rule}
Table~\ref{tab:mlpalloc} makes the failure concrete, and the oracle's assignments are the substantive finding of this section. At every fractional budget it returns a \emph{flat} allocation split by matrix role: $\texttt{c\_fc}$ at $b{+}1$ and $\texttt{mlp.c\_proj}$ at $b{-}1$, with no depth variation whatever (3/2 at 2.5 bits, 4/3 at 3.5, 8/4 at 6). C1 instead spends its budget along depth --- at the 4-bit budget it robs $\texttt{c\_proj}$ in blocks 0--4 down to 3 bits to fund an 8-bit $\texttt{c\_proj}$ in block 11, and loses $1.23$ perplexity against leaving everything at 4. A plausible mechanism is that both factors of $\Omega_i(b) = \mathrm{Tr}(H_i)\,\|Q_b(W_i)-W_i\|_F^2$ scale with magnitude while GPTQ's per-group scales already normalize for it, so the criterion reads activation scale --- which grows with depth in a residual stream --- as curvature. The measured block reconstruction losses track this directly, rising from $0.033$ at block 0 to $0.113$ at block 10. A scale-normalized trace $\mathrm{Tr}(H_i)/\overline{\mathrm{diag}(H_i)}$ is the obvious test and was not run.

\paragraph{Fine-tuning is inert in the deep layers}
Section~\ref{sec:ft} reports a flip rate of $0.12$--$0.18$ throughout and reads it as evidence the stage reaches deployed weights. The MLP run does not reproduce this. Pooling the 20 fine-tuned sweep arms, the flip rate is exactly zero in blocks 9, 10 and 11 in $15$, $12$ and $17$ arms respectively, against zero occurrences in blocks 0--3; the reconstruction loss is unchanged to six decimals in those blocks, so the commit gate of Section~\ref{sec:ft} rejected every one of 200 steps. The signature is the one Section~\ref{sec:ft} already attributes to a fixed absolute learning rate, and the deep blocks are precisely those carrying the largest reconstruction error, so the stage fails where it is most needed. The aggregate consequence is that fine-tuning is worth $\approx1\%$ here against $1$--$2\%$ on QKV, it hurts at two budgets rather than one, and the depth-restricted control lands at $26.12$ between GPTQ-only $26.21$ and full-depth $25.93$ --- a much narrower band than the QKV analogue. We report the Objective-3 results on this sublayer as unresolved rather than negative.

\paragraph{What this says about the central claim}
Section~\ref{sec:ft} establishes that the KL attention-map term supplies $88\%$ of the fine-tuning gain, and Fig.~\ref{fig:ft}(b) that removing it is harmful at every budget. The MLP port is the limiting case of that ablation: the term is not down-weighted but structurally absent, and both halves of the pipeline degrade together --- the criterion falls behind uniform, and refinement falls to a percent. The honest reading is that what transfers from this work is the attention-specific machinery, not a general-purpose mixed-precision recipe, and that a sublayer without a distributional target to distill against needs a different objective rather than the same one with a term deleted.

\paragraph{What is not measured here}
A single seed was run for Table~\ref{tab:mlp} ($\lambda_{\text{KL}}=0$, seed 42, full WikiText-2 test split, $242$ minutes over 44 quantize-and-evaluate passes), so L1 and L5 apply unchanged. Integer budgets carry no information about allocation in this setting: at 3.0 and 4.0 bits every criterion returns a flat assignment and the four columns of Table~\ref{tab:mlp} coincide exactly, as the budget arithmetic of Section~\ref{sec:alloc} requires.

\paragraph{A second, C4-calibrated run does instrument $\varepsilon^W$, $\varepsilon^A$ and $\rho$}
A separate MLP sweep, calibrated on C4 rather than WikiText-2 and following Run B's protocol (four modes, five budgets, scored on both corpora), defines $\varepsilon^{A}$ as the relative Frobenius error of the MLP block's own output --- the direct analogue of the attention-output $\varepsilon^{A}$ of Section~\ref{sec:metrics}, with $\texttt{mlp.c\_proj}$'s input playing the role QKV's output plays there --- and reports $\rho=\varepsilon^{A}/\varepsilon^{W}$ alongside it (Table~\ref{tab:mlprunb}).

\begin{table*}[!t]
\centering
\caption{GPT-2 MLP, C4 calibration: four pipeline modes, five budgets, evaluated on both corpora, with the amplification factor $\rho=\varepsilon^{A}/\varepsilon^{W}$ reported per corpus ($\varepsilon^{W}$ is shared). Controls: WikiText-2 24.357/0.4148; C4 32.770/0.3736. Not directly comparable to Table~\ref{tab:mlp}, which is calibrated on WikiText-2.}
\label{tab:mlprunb}
\tiny
\setlength{\tabcolsep}{3pt}
\begin{tabular}{l c cccc cccc c}
\toprule
& & \multicolumn{4}{c}{WikiText-2 (cross-corpus)} & \multicolumn{4}{c}{C4 (in-domain)} & \\
\cmidrule(lr){3-6}\cmidrule(lr){7-10}
Mode & bits & PPL & Acc & $\varepsilon^{A}$ & $\rho$ & PPL & Acc & $\varepsilon^{A}$ & $\rho$ & $\varepsilon^{W}$ \\
\midrule
uniform & 3   & 54.152  & 0.3241 & 0.4749 & 1.332 & 61.630  & 0.3035 & 0.4632 & 1.299 & 0.3565 \\
uniform & 4   & 26.423  & 0.4053 & 0.2112 & 1.349 & 34.765  & 0.3652 & 0.2068 & 1.321 & 0.1566 \\
uniform & 6   & 24.420  & 0.4149 & 0.0483 & 1.338 & 32.648  & 0.3734 & 0.0472 & 1.307 & 0.0361 \\
\midrule
jab & 3   & 600.798 & 0.1219 & 0.4650 & 1.192 & 574.850 & 0.1188 & 0.4457 & 1.143 & 0.3900 \\
jab & 3.5 & 46.236  & 0.3421 & 0.3292 & 1.273 & 54.889  & 0.3177 & 0.3237 & 1.251 & 0.2587 \\
jab & 4   & 26.423  & 0.4053 & 0.2112 & 1.349 & 34.765  & 0.3652 & 0.2068 & 1.321 & 0.1566 \\
jab & 4.5 & 26.237  & 0.4065 & 0.2009 & 1.387 & 34.621  & 0.3661 & 0.1959 & 1.353 & 0.1448 \\
jab & 6   & 25.499  & 0.4103 & 0.1113 & 1.305 & 33.779  & 0.3690 & 0.1094 & 1.283 & 0.0853 \\
\midrule
joint & 3 & 48.121 & 0.3331 & 0.4199 & 1.038 & 58.676 & 0.3044 & 0.4021 & 0.994 & 0.4045 \\
joint & 4 & 26.055 & 0.4054 & 0.1951 & 1.123 & 34.411 & 0.3665 & 0.1873 & 1.078 & 0.1738 \\
joint & 6 & 24.434 & 0.4143 & 0.0469 & 1.206 & 32.731 & 0.3737 & 0.0452 & 1.162 & 0.0389 \\
\midrule
adaptive\_joint & 3   & 157.931 & 0.2067 & 0.4203 & 0.904 & 152.766 & 0.2195 & 0.4015 & 0.863 & 0.4651 \\
adaptive\_joint & 3.5 & 40.772  & 0.3474 & 0.2864 & 0.965 & 51.422  & 0.3197 & 0.2742 & 0.924 & 0.2967 \\
adaptive\_joint & 4   & 26.055  & 0.4054 & 0.1951 & 1.123 & 34.411  & 0.3665 & 0.1873 & 1.078 & 0.1738 \\
adaptive\_joint & 4.5 & 25.920  & 0.4062 & 0.1847 & 1.140 & 34.283  & 0.3669 & 0.1764 & 1.089 & 0.1620 \\
adaptive\_joint & 6   & 25.349  & 0.4099 & 0.0965 & 0.959 & 33.503  & 0.3689 & 0.0914 & 0.909 & 0.1006 \\
\bottomrule
\end{tabular}
\end{table*}

\paragraph{$\rho$ separates GPTQ-only from fine-tuned arms less cleanly than on QKV}
GPTQ-only allocation (uniform, jab) has $\rho>1$ at every budget on both corpora, as on QKV. But the two fine-tuned modes do not fall cleanly below $\rho=1$: \texttt{joint} stays above $1$ except at the tightest budget, while \texttt{adaptive\_joint} dips below $1$ at 3, 3.5 and 6 bits but rises back above it at 4 and 4.5, tracking the flat-assignment coincidence with \texttt{jab} noted in Section~\ref{sec:mlp} rather than the fine-tuning stage itself. The clean, crossing-free separation of Fig.~\ref{fig:err}(a,b) is therefore a QKV-specific regularity, not a general property of $\rho$.

\paragraph{The 3-bit allocator failure is far more severe under C4 calibration}
Table~\ref{tab:mlp}'s WikiText-2-calibrated \texttt{jab} arm loses to uniform at 3 bits by $2.25\times$ ($120.70$ vs.\ $53.71$); under C4 calibration the same criterion at the same budget loses by $11.1\times$ ($600.798$ vs.\ $54.152$), with \texttt{adaptive\_joint} similarly worse ($157.931$ vs.\ Table~\ref{tab:mlp}'s finetuned $94.49$). The failure mode is the same one-layer 2-bit collapse described in Section~\ref{sec:mlp}, but its severity is calibration-corpus dependent in a way the QKV sweep's $r=0.9995$ trace correlation (Section~\ref{sec:sens}) did not predict.
\section{Full-Model Quantization: Extended Analysis}
\label{sec:fullmodel}

Sections~\ref{sec:sens}--\ref{sec:cross} quantize the QKV projection and Section~\ref{sec:mlp}
the MLP, each in isolation. We now compose them: every linear module of every Mistral-7B
decoder layer is quantized under a single budget, covering $6.98$B parameters ($96.4\%$ of the
model, 224 allocation units). This is the regime practitioners deploy, and it tests three
claims that the sublayer studies could only suggest: whether a sensitivity criterion helps once
units are heterogeneous, whether block-local refinement remains safe, and whether the $2$-bit
allocation pathology of Section~\ref{sec:allocres} is specific to GPT-2.

\subsection{Setting}

\paragraph{Implementation}
Four Hessians suffice per block: $q,k,v$ share the input $\mathrm{RMSNorm}_1(h)$ and
\texttt{gate},\,\texttt{up} share $\mathrm{RMSNorm}_2(h)$. They are collected in forward order
$qkv \to o \to \{\texttt{gate},\texttt{up}\} \to \texttt{down}$, each after the preceding
units are quantized, extending within the block the sequential property of
Section~\ref{sec:gptq}. Weight layout is an explicit per-architecture flag, since
\texttt{nn.Linear} stores $(d_{\mathrm{out}},d_{\mathrm{in}})$, the transpose of GPT-2's
\texttt{Conv1D}, and a silent transpose would quantize along the wrong axis. Unlike the
streaming build of Section~\ref{sec:setup}, this run solves GPTQ in float64 throughout with
per-block offloading on a single 96\,GB GPU, so the float32-solve caveat of
Section~\ref{sec:gptq} does not apply. A hand-written block forward reproducing RoPE, RMSNorm,
GQA head repetition and SwiGLU is validated against \texttt{MistralDecoderLayer} before any
arm runs. Evaluation uses 32 WikiText-2 windows of 512 tokens (fp16 control $6.643$), one
seed, and 64 Hessian batches.

\begin{figure}[t]
\centering
\includegraphics[width=0.92\textwidth]{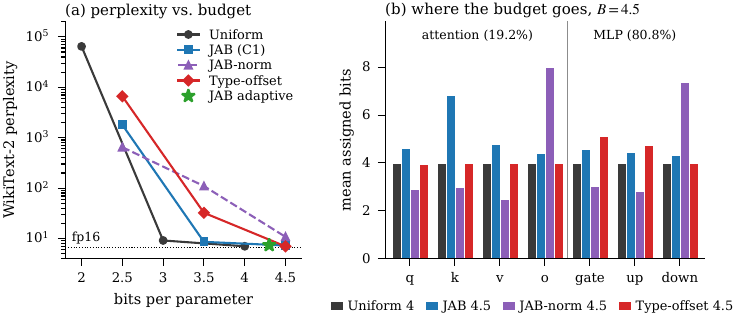}
\caption{Full-model quantization of Mistral-7B. \textbf{(a)} Perplexity against bits per
parameter, GPTQ only. At $B=4.5$ Type-offset is best; the adaptive JAB arm ($\star$) spends
$4.3$ bits per parameter yet scores worse than uniform $4$-bit. \textbf{(b)} Mean assigned
bits per unit kind at $B=4.5$. Type-offset is flat within a role and offset between roles; JAB
over-buys \texttt{k\_proj}, which grouped-query attention makes cheap; JAB-norm assigns
$8$ bits to every \texttt{o\_proj} while starving $q,k,v$ to $2$--$3$ bits.}
\label{fig:fullmodel}
\end{figure}

\subsection{Layer Role Outperforms Curvature}
\label{sec:roleresult}

\paragraph{A seven-integer prior beats the Hessian criterion at matched budget}
At $B=4.5$ (Table~\ref{tab:fullmodel}), Type-offset reaches $6.970$ against JAB's $7.250$ and
JAB-norm's $10.77$ --- all three at exactly $4.500$ realized bits per parameter --- while
computing no Hessian, no Hutchinson probe and no scoring pass. This is the second architecture
in which role-based allocation matches or beats the learned criterion: the GPT-2 MLP oracle
found the same flat, role-split structure under GELU (Section~\ref{sec:mlp}), and here it
prevails under SwiGLU with grouped-query attention and $56\times$ more parameters.

The sensitivity-driven allocator also fails a weaker test. At $B=4.3$ the adaptive JAB arm
reaches $7.373$, worse than uniform $4$-bit ($6.998$) while spending $7.5\%$ more bits.
Uniform $4$-bit remains within $0.03$ of the best mixed-precision arm at $B=4.5$, so in this
regime no allocator we tested improves on uniform quantization per bit spent; what
distinguishes them is how much each loses.

\paragraph{Why the criterion cannot find role structure}
The JAB trace is computed once per block over the concatenated parameter vector, and is
therefore identical for all seven units of a layer (e.g.\ $1478.6$ for every unit of
block~5). Between units of one block, the criterion can discriminate only through the
perturbation factor of Eq.~\ref{eq:omega}. Its dominant signal is instead a depth gradient
spanning $381\times$ ($59.2$ at block~0 to $22{,}575$ at block~31) --- the one axis on which
the winning allocation has no structure at all. Fig.~\ref{fig:fullmodel}(b) shows the
consequence: JAB's largest purchases are GQA-cheap \texttt{k\_proj} units rather than any role.
Per-unit traces would remove this limitation at four double-backward passes per block.

\paragraph{Scale normalization trades one bias for another}
Section~\ref{sec:mlp} hypothesized that the criterion reads activation magnitude as curvature
and proposed normalizing the trace by $\overline{\mathrm{diag}(H_i)}$. We tested this
(JAB-norm). It is worse than JAB at both usable budgets ($10.77$ vs.\ $7.250$ at $B=4.5$;
$109.6$ vs.\ $8.570$ at $B=3.5$). Dividing by the mean Hessian diagonal rewards units whose
input spectrum is flat, and \texttt{o\_proj}, fed by the merged attention output, is far flatter
than the post-RMSNorm stream feeding $q,k,v$; JAB-norm consequently assigns $8.0$ mean bits to
\texttt{o\_proj} and $7.4$ to \texttt{down\_proj} while leaving $q,k,v$ at $2.5$--$3.0$. The
normalizer exchanges a magnitude bias for a conditioning bias.

\paragraph{The offset sign matters, and is the main open parameter}
Type-offset degrades sharply at $B=3.5$ ($32.33$). With $\Delta_{\texttt{down}}=-1$, the
allocation places $30$ of $32$ \texttt{down\_proj} matrices --- $25.2\%$ of the quantized
weights --- at $2$ bits, the width that destroys uniform quantization. At $B=4.5$ the grid
$\mathcal{B}$ absorbs the offset ($4-1$ and $4+1$ round back to $4$) and the budget is instead
spent promoting individual \texttt{gate} and \texttt{up} units to $8$ bits. SwiGLU's
\texttt{down\_proj} receives $\mathrm{silu}(\texttt{gate}(x))\odot\texttt{up}(x)$, a
heavier-tailed input than the post-GELU activation of its GPT-2 counterpart, so the transposed
negative offset is plausibly the wrong sign. Because each candidate offset vector costs a single
evaluation, fitting $\Delta$ is inexpensive; we report the untuned vector, so the $B=4.5$ result
is a lower bound on what the family achieves.

\subsection{Block-Local Objectives Can Diverge from End-to-End Quality}
\label{sec:proxygap}

Section~\ref{sec:ft} identified a gap between the fine-tuning objective and perplexity that the
commit gate cannot detect, of under $1\%$ on QKV. Under full coverage the same mechanism
produces a $32\times$ degradation, and a depth-restricted control localizes it to a single
block (Table~\ref{tab:ftcontrol}).

\begin{table}[t]
\centering
\caption{Joint fine-tuning under full coverage. ``Changed'' counts blocks whose quantized weights
differ from the GPTQ warm start (flip rate $>0$, Eq.~\ref{eq:flip}); ``Block-0 loss'' is that
block's own objective before and after refinement. Restricting refinement to blocks 0--3 reproduces
the full-depth result exactly.}
\label{tab:ftcontrol}
\small
\setlength{\tabcolsep}{5pt}
\begin{tabular}{@{}l c c c r@{}}
\toprule
Allocation, refinement & Blocks refined & Changed & Block-0 loss & PPL \\
\midrule
Adaptive $B{=}4.3$, GPTQ only           & --        & --        & --                          & 7.373 \\
\quad + joint FT                        & 0--31     & 1 (0)     & $1.85\to0.41\times10^{-3}$  & \textbf{239.52} \\
\quad + joint FT                        & 0--3      & 1 (0)     & $1.85\to0.41\times10^{-3}$  & \textbf{239.52} \\
\quad + joint FT, $\lambda_{\mathrm{KL}}{=}0$ & 0--31 & 1 (0)   & MSE only                    & 52.07 \\
\addlinespace
Uniform $4$, GPTQ only                  & --        & --        & --                          & 6.998 \\
\quad + joint FT                        & 0--31     & 3 (0,3,4) & $1.0\to0.9\times10^{-5}$    & 11.39 \\
\bottomrule
\end{tabular}
\end{table}

\paragraph{The measurement}
Refining the adaptive allocation raises perplexity from $7.373$ to $239.52$. Refining only
blocks 0--3 yields $239.52$ as well, identical to four decimals together with every
weight- and activation-error metric. The flip-rate diagnostic explains the coincidence: in both
runs block~0 is the only block whose quantized weights change (flip rate $0.043$); in the other
31 blocks every weight returns to its GPTQ grid point. Within block~0, the refinement
\emph{succeeded} on its own terms: its objective fell $4.6\times$, from
$1.85\times10^{-3}$ to $0.41\times10^{-3}$, and the held-out commit gate accepted it. One
refined block in 32 multiplied end-to-end perplexity by $32.5$.

\paragraph{Mechanism}
Two conditions coincide at block~0. The adaptive allocator assigns it $2$ bits on six of its
seven units, giving a mean relative weight error $\varepsilon^W=1.015$ ($1.42$ for
\texttt{q\_proj}): the quantized block lies farther from its float weights than the zero matrix
does. And block~0's output feeds all 31 blocks above it, so its residual is the least local
quantity in the network. A $200$-step descent that reduces this block's objective on eight
calibration batches, from a starting point already beyond the zero-error reference, fits noise
with maximal downstream leverage. Severity tracks grid coarseness as this account predicts:
on the uniform $4$-bit allocation the same stage changes three blocks and degrades perplexity
by $1.6\times$ rather than $32\times$.

\paragraph{Implication beyond JAB}
Nothing in this mechanism is specific to $\Ljab$. Any post-training
pipeline that refines blocks against block-local targets and validates on held-out
calibration data admits the same failure, and the local measurement alone places no bound on
the global damage. A certified improvement in a block's own reconstruction objective is
therefore not evidence of end-to-end improvement. Reporting a global metric per refinement
stage, together with a per-block flip rate, is the minimum needed to detect it.

\paragraph{The KL term amplifies an ill-posed refinement}
On QKV the attention-map term supplies $88\%$ of the fine-tuning gain (Section~\ref{sec:ft}).
Under full coverage, removing it reduces the damage from $239.52$ to $52.07$, with the block-0
flip rate falling from $0.043$ to $0.034$. Both remain far worse than no refinement, so this
compares two degraded configurations; it shows that the KL term drives the stage harder, which
helps when the stage is well posed and harms when it is not. Its effect is a property of the
regime, not of the objective.

\subsection{The 2-Bit Pathology Is Not Specific to GPT-2}
\label{sec:twobit}

Section~\ref{sec:allocres} proved that GPT-2's $3.0$-bit assignment must contain a sub-$3$-bit
block, and recorded as an open discrepancy that no comparable collapse appeared on Mistral.
Full coverage resolves it. The adaptive allocation at $B=4.3$ places its only sub-$3$-bit units
in block~0 --- six of seven at $2$ bits, $\varepsilon^W=1.015$, the only block with
$\varepsilon^W>1$ --- and the arm loses to uniform while spending more. The earlier absence was
a matter of scope rather than of depth or grouped-query attention: with only QKV quantized, a
single $2$-bit projection is $1/128$ of the quantized weights; with all seven modules under
Eq.~\ref{eq:pwbudget}, the allocator can concentrate low widths on one block, and the additive
objective of Eq.~\ref{eq:mckp} still cannot represent that perplexity is convex and unbounded
in the low-bit direction. The floor $b_i\geq3$ proposed in Section~\ref{sec:conclusion} removes
every sub-$3$-bit unit from this assignment as it does from GPT-2's; whether it recovers
uniform-level quality at $7$B is untested.

\paragraph{Scope}
All results in this section come from one seed. The $0.28$ separating Type-offset from JAB at
$B=4.5$ and the effects of Section~\ref{sec:proxygap} are large relative to that limitation;
the $0.03$ separating Type-offset at $B=4.5$ from uniform $4$-bit is not, and the two differ in
budget. Perplexities are exact within this run but computed on a 32-window subset. Beyond
block~0, the fine-tuning stage changed no weights in 31 of 32 blocks, the mis-scaled step
signature of Section~\ref{sec:ste}; refinement is therefore under-exercised here independently
of the divergence above, and Type-offset was evaluated without it.

\subsection{GPT-2: QKV\texorpdfstring{$+$}{+}MLP, Full WikiText-2 Table}

\paragraph{Role beats curvature once both sublayers share the budget.} At the aggressive fractional
budgets, Type-offset needs no scoring pass yet leads: $64.896$ at $3.5$ bits and $28.211$ at $4.5$,
against JAB+prop's $67.337$ and $28.694$ (Table~\ref{tab:gpt2full}). JAB's attention-defined
sensitivity over-invests in QKV, a quarter of the quantized weights, and under-funds the MLP, which
now dominates the parameter count; Type-offset's flat role split needs no such trade-off. Only near
saturation, at 6 bits, does uniform win ($24.546$): perplexity is nearly flat there, so a mixed
4-/8-bit average cannot beat a flat assignment, and Type-offset still shifts bits toward the MLP
while QKV lags.

\paragraph{Attention error stops tracking perplexity.} On QKV alone, attention-output error
$\varepsilon^A$ tracked perplexity closely. Once the MLP shares the budget this relationship breaks:
$\varepsilon^A$ sees only QKV and is blind to the MLP, which dominates the quantized weights.
Type-offset has the worst $\varepsilon^A$ at $3.5$ and $4.5$ bits yet the best perplexity, and
accuracy tracks perplexity rather than $\varepsilon^A$ throughout (Table~\ref{tab:gpt2full}).
Sublayer damage also compounds: QKV and MLP errors multiply once both are quantized, so at 3 bits the
composed model is substantially farther from fp32 than either sublayer alone would suggest ---
consistent with each MLP being calibrated on an already-quantized attention block.

\paragraph{Block-local fine-tuning wins locally, loses globally.} Adding the joint STE stage to
uniform assignments raises perplexity at every budget --- $16.35\times$ at 3 bits --- even as
block-reconstruction loss falls several-fold and the held-out gate commits nearly every block. The
stage minimizes block-output MSE while perplexity is a causal, end-to-end quantity that no block-local loss can see; the
held-out gate certifies exactly the updates responsible for the damage. No fine-tuned arm beats its
GPTQ-only counterpart: GPTQ-only Type-offset at $4.5$ bits ($28.211$) beats fine-tuned uniform 4-bit
($30.65$).

\label{sec:gpt2wiki}

\begin{table}[t]
\centering
\caption{GPT-2 with QKV and MLP quantized together (C4 calibration, GPTQ-only unless noted):
WikiText-2 perplexity (fp32 control $24.357$), accuracy, attention error $\varepsilon^A$, weight
error $\varepsilon^W$, and $\rho = \varepsilon^A/\varepsilon^W$. C4 (in-domain, control $32.770$)
gives the same ranking; full C4 numbers are in Appendix~\ref{sec:gpt2c4} (Table~\ref{tab:gpt2c4}).
Best per budget in bold. }
\label{tab:gpt2full}
\small
\setlength{\tabcolsep}{5pt}
\begin{tabular}{@{}l c c c c c c@{}}
\toprule
Allocator & $B$ & PPL & Acc & $\varepsilon^A$ & $\varepsilon^W$ & $\rho$ \\
\midrule
Uniform              & 3.0 & 141.821 & 0.232 & 0.414 & 0.464 & 0.892 \\
JAB                   & 3.5 & 75.009  & 0.296 & 0.267 & 0.383 & 0.697 \\
JAB+prop              & 3.5 & 67.337  & 0.307 & 0.320 & 0.379 & 0.845 \\
Type-offset  & 3.5 & \textbf{64.896} & \textbf{0.310} & 0.421 & 0.303 & 1.390 \\
All (flat)            & 4.0 & 29.308  & 0.393 & 0.238 & 0.210 & 1.137 \\
JAB                   & 4.5 & 29.041  & 0.394 & 0.218 & 0.188 & 1.159 \\
JAB+prop              & 4.5 & 28.694  & \textbf{0.397} & 0.194 & 0.210 & 0.924 \\
Type-offset  & 4.5 & \textbf{28.211} & 0.395 & 0.237 & 0.144 & 1.645 \\
Uniform               & 6.0 & \textbf{24.546} & \textbf{0.413} & 0.070 & 0.048 & 1.449 \\
JAB                   & 6.0 & 26.868  & 0.404 & 0.071 & 0.152 & 0.470 \\
JAB+prop              & 6.0 & 25.899  & 0.407 & 0.121 & 0.141 & 0.859 \\
Type-offset  & 6.0 & 27.441  & 0.400 & 0.245 & 0.128 & 1.914 \\
\bottomrule
\end{tabular}
\end{table}

\subsection{GPT-2: C4 Results for QKV+MLP}
\label{sec:gpt2c4}

Section~\ref{sec:g-full} reports the GPT-2 QKV+MLP results on WikiText-2 (Table~\ref{tab:gpt2full})
and states that C4, the in-domain calibration corpus, gives the same ranking. Table~\ref{tab:gpt2c4}
gives those C4 numbers in full. Type-offset leads at $3.5$ bits ($78.239$) and $4.5$ bits
($37.045$), uniform wins at $6.0$ bits ($32.935$), and attention error $\varepsilon^A$ is again
uninformative --- Type-offset has the worst $\varepsilon^A$ at $3.5$ and $4.5$ bits while still
leading on perplexity and accuracy --- reproducing every qualitative claim of
Section~\ref{sec:g-full} on a disjoint, in-domain corpus.

\begin{table}[t]
\centering
\caption{GPT-2 with QKV and MLP quantized together, evaluated on C4 (in-domain; fp32 control
$32.770$), matching the WikiText-2 results of Table~\ref{tab:gpt2full}. Same allocators, budgets
$B$ (bits/parameter), and metrics. Best per budget in bold.}
\label{tab:gpt2c4}
\small
\setlength{\tabcolsep}{5pt}
\begin{tabular}{@{}l c c c c c c@{}}
\toprule
Allocator & $B$ & PPL & Acc & $\varepsilon^A$ & $\varepsilon^W$ & $\rho$ \\
\midrule
Uniform              & 3.0 & 155.246 & 0.220 & 0.400 & 0.448 & 0.894 \\
JAB                   & 3.5 & 83.311  & 0.279 & 0.261 & 0.367 & 0.711 \\
JAB+prop              & 3.5 & 81.408  & 0.280 & 0.316 & 0.371 & 0.852 \\
Type-offset  & 3.5 & \textbf{78.239} & \textbf{0.287} & 0.421 & 0.303 & 1.390 \\
All (flat)            & 4.0 & 39.053  & 0.352 & 0.237 & 0.205 & 1.160 \\
JAB                   & 4.5 & 38.626  & 0.355 & 0.219 & 0.183 & 1.195 \\
JAB+prop              & 4.5 & 38.132  & 0.356 & 0.193 & 0.206 & 0.936 \\
Type-offset  & 4.5 & \textbf{37.045} & \textbf{0.357} & 0.237 & 0.144 & 1.642 \\
Uniform               & 6.0 & \textbf{32.935} & \textbf{0.373} & 0.070 & 0.047 & 1.496 \\
JAB                   & 6.0 & 35.725  & 0.364 & 0.068 & 0.149 & 0.458 \\
JAB+prop              & 6.0 & 34.217  & 0.367 & 0.120 & 0.138 & 0.871 \\
Type-offset  & 6.0 & 36.334  & 0.361 & 0.248 & 0.128 & 1.939 \\
\bottomrule
\end{tabular}
\end{table}

\subsection{Per-Unit Traces and the 3-Bit Floor}
\label{sec:perunit}

\paragraph{Estimator}
With a Rademacher probe $v$ over the concatenated seven-unit vector and $Hv$ the existing
Hessian-vector product, let $S_u$ be unit $u$'s coordinate range and
$t_u = \sum_{j\in S_u} v_j (Hv)_j$. Since $\mathbb{E}[v_jv_k]=\delta_{jk}$,
$\mathbb{E}[t_u]=\sum_{j\in S_u} H_{jj} = \operatorname{Tr}(H_{uu})$: the partial sums are unbiased
for the diagonal-block traces, and $\sum_u t_u$ is exactly the pooled estimate used elsewhere, which
we assert at run time. The cross-block terms cancel in the pooled sum but not per unit, so we raise
the probe count from 10 to 30; the scoring pass runs once per run, not once per arm.

\paragraph{JAB-norm}
Normalizing by $\overline{\mathrm{diag}(H_i)}$ favors units with flatter input spectra: without the
floor it allocates substantially more bits to $o_{\mathrm{proj}}$ while assigning only $2$--$3$ bits
to $q,k,v$, and it performs worse than JAB at both usable budgets, in that run and in the floored
run of Table~\ref{tab:floor}.

\paragraph{The trace mass is concentrated on Q and K}
Median per-unit traces across the 32 blocks are $2081.6$ ($q$), $1245.6$ ($k$), $365.5$
(\texttt{down}), $211.0$ ($v$), $160.2$ (\texttt{up}), $146.0$ ($o$) and $93.3$ (\texttt{gate}).
Attention thus receives $85\%$ of the median trace mass while holding $19.2\%$ of the quantized
weights. This is a property of the objective rather than of the estimator: $\Ljab$ is evaluated on
the attention output and post-softmax map, so perturbing an MLP matrix affects it only through the
block output term.

\paragraph{Within-block discrimination}
Where the pooled trace gave one value for all seven units of a block, the per-unit traces in block 5
range from $878.9$ for $q_{\mathrm{proj}}$ to $7.5$ for $\mathrm{gate}_{\mathrm{proj}}$, a
$117\times$ spread within a single block.

\paragraph{Allocations at $B=4.5$}
Mean assigned bits per unit kind, JAB versus Type-offset:
$q$ $7.47$ vs.\ $3.97$, $k$ $7.38$ vs.\ $4.00$, $v$ $4.06$ vs.\ $4.00$, $o$ $3.88$ vs.\ $3.97$,
\texttt{gate} $3.94$ vs.\ $5.12$, \texttt{up} $4.09$ vs.\ $4.75$, \texttt{down} $4.62$ vs.\ $4.00$.
JAB spends its surplus lifting $q$ and $k$ to nearly 8 bits; Type-offset spends it on
\texttt{gate} and \texttt{up}. Both realize $4.500$ bits per parameter exactly, and the second
allocation scores $0.225$ perplexity lower.

\paragraph{Protocol and comparability}
One seed, floor $b_i\ge3$ so the candidate set is $\{3,4,8,16\}$, 30 Hutchinson probes, 128 Hessian
batches, 32 WikiText-2 windows of 512 tokens. Budgets below the floor are infeasible and were
skipped, so this run has no $B=2.5$ column. Because the Hessian batch count differs from the run of
Table~\ref{tab:fullmodel} (128 against 64), absolute perplexities are not interchangeable between
the two tables --- uniform 4-bit moves from $6.998$ to $6.938$ --- and only within-table comparisons
are exact.

\section{Limitations in Full}
\label{sec:limits}
\begin{enumerate}[label=L\arabic*.,leftmargin=*,itemsep=1pt,topsep=2pt]
\item \textit{Single-draw trace estimates.} Ten Hutchinson probes on 2 batches, no repeated-seed variance report; in particular the block-0/block-1 trace gap driving the 3.0-bit failure is not shown to exceed sampling noise, and neither is the $4.5\times$ MLP depth gradient of Section~\ref{sec:mlp}.
\item \textit{Scope.} Sections~\ref{sec:sens}--\ref{sec:cross} quantize the QKV projection and Section~\ref{sec:mlp} the MLP; Section~\ref{sec:fullmodel} composes all seven Mistral-7B modules ($96.4\%$ of parameters). The composition was run on Mistral only, so the sublayer interaction is unmeasured on GPT-2. Embeddings and the output head are never quantized, capping end-to-end compression at $1.66\times$ on GPT-2 at 4 bits.
\item \textit{Greedy allocation is unverified at full scale.} Its optimality gap was never measured, exhaustive enumeration being intractable, so how much of the 3.0-bit failure is the criterion and how much the solver is unknown; that greedy and MCKP/Pareto agree exactly on Mistral is suggestive, not a proof.
\item \textit{The joint-versus-separate gap is unmeasured.} Section~\ref{sec:joint} shows the cross-blocks $\Gamma$ are structurally non-zero and predicts what a per-matrix baseline would look like, but no Q/K/V-separate arm was run and $\|\Gamma\|_F/\|H_{\mathcal{L}}\|_F$ was never estimated, so the size of the effect on real blocks is unknown.
\item \textit{The MLP fine-tuning result is confounded.} Section~\ref{sec:mlp} records a zero flip rate in the deepest blocks, the signature Section~\ref{sec:ft} attributes to a mis-scaled learning rate. Until the grid step is confirmed to be computed per unit rather than pooled across units of differing input dimension, the weak Objective-3 result there cannot be separated from an optimizer defect.
\item \textit{Single seed; the Mistral half is convergent evidence, not a replicate.} One seed per pairing, a re-implementation not checked against the full GPT-2 validation suite, a float32 rather than float64 solve (Section~\ref{sec:gptq}), and a different $\varepsilon^W,\varepsilon^A$ normalization.
\end{enumerate}

\end{document}